\pdfoutput=1
\documentclass{article}
\usepackage{iclr2027_conference,times}

\usepackage[utf8]{inputenc}
\usepackage[T1]{fontenc}
\usepackage{amsmath,amsfonts,amssymb,bm}
\usepackage{graphicx}
\usepackage{subcaption}
\usepackage{booktabs}
\usepackage{listings}
\usepackage{multirow}
\usepackage{placeins}
\usepackage{microtype}
\usepackage{xcolor}
\usepackage{tikz}
\usetikzlibrary{positioning,fit,calc,backgrounds}
\usepackage{hyperref}
\usepackage{url}

\newcommand{\benchmark}{FaithAlign}
\newcommand{\contextgap}{FaithAlign-Gap}
\newcommand{\numModels}{8}

\title{Aligning Faithfulness for Theory-Scale \\ Autoformalization}

\author{%
  Yuming Feng$^{1}$\quad
  Tadd Mao$^{2}$\quad
  Dhruva Arekar$^{1}$\quad
  Tianjun Zhong$^{1}$\\
  \bfseries Jiani Huang$^{3}$\quad
  Xujie Si$^{2}$\quad
  Ziyang Li$^{1}$\\
  {\normalfont $^{1}$Johns Hopkins University\quad
  $^{2}$University of Toronto\quad
  $^{3}$University of Pennsylvania}\\
  {\normalfont\ttfamily\fontsize{8}{10}\selectfont yfeng97@cs.jhu.edu, tadd.mao@mail.utoronto.ca, darekar1@jhu.edu,}\\
  {\normalfont\ttfamily\fontsize{8}{10}\selectfont tzhong4@jhu.edu, jianih@seas.upenn.edu, six@cs.toronto.edu, ziyang@cs.jhu.edu}
}

\iclrfinalcopy

\begin{document}

\maketitle

\begin{abstract}

Autoformalization (AF) requires more than compilation: formal statements must preserve the meaning of their informal sources, a property known as \textit{faithfulness}.
An emerging frontier is the automated evaluation of this property, yet most existing methods operate at the statement level and produce whole-item verdicts, failing to reveal how individual expressions correspond or which contextual interpretations support the judgment.
This omission is consequential because AF is inherently theory-scale: target statements derive their meaning from surrounding mathematical material and unstated conventions.
We formulate faithfulness evaluation as a context-aware \textit{align-then-label} problem: identifying phrase-level correspondences between informal and formal statements and assigning a faithfulness label to each under explicitly specified contextual interpretations and assumptions.
We also introduce \benchmark{}, the first theory-scale faithfulness benchmark comprising 460 items and 4,982 annotated phrase-level links, alongside \contextgap{}, a companion set of 146 items with informal context and assumptions withheld.
Each \benchmark{} item includes the target statement pair, relevant context, formal dependencies, and, where needed, assumptions specifying one given interpretation.
Across \numModels{} models, GPT-6 Astra achieves the highest defect F1 at 47.8\% and identifies all defects in only 4.5\% of items with two or more defects.
On \contextgap{}, Astra identifies only 28.2\% of undecidable items as such.
These results show the difficulty of recovering context-dependent correspondences, establishing \benchmark{} as a testbed for fine-grained, context-aware faithfulness evaluation.

\end{abstract}

\section{Introduction}
\label{sec:introduction}

Autoformalization aims to translate mathematical writing into formal statements and definitions while preserving its meaning---a property known as \emph{faithfulness}.
This task is \textit{theory-scale}: formalizing a mathematical definition requires tracing its dependencies on other definitions, standing assumptions, and conventions \citep{min2026theory,feng2026lcsbench}.
A formalization may look faithful in isolation but narrow the domain, omit an assumption, or misstate a definition in its original context, causing cascading failures in theory-scale autoformalization.
Assessing faithfulness therefore requires identifying each discrepancy and tracing the dependencies behind it.
When context is missing, the available information may support different interpretations.
In such cases, faithfulness is \emph{undecidable}.

Existing evaluations largely reduce faithfulness to a scalar score or binary verdict for a pair of informal and formal statements \citep{lu2025formalalign,poiroux2025reliable,patel2026mathatlas}.
Such outputs do not map the surrounding informal context to formal dependencies or show whether all defects have been localized.
AriaScorer grounds binary judgments in retrieved library definitions but does not measure coverage of contextual correspondences~\citep{wang2026aria}.
FormalRx localizes problematic formal code without linking it to informal expressions and their dependencies~\citep{wang2026formalrx}.
The binary labels used by MA-Align and FormalRx cannot represent \emph{undecidable} cases where context is underspecified \citep{patel2026mathatlas,wang2026formalrx}.
A definite verdict thus assumes an interpretation unsupported by the supplied material.

We introduce a novel context-aware \emph{align-then-label} problem to address these gaps.
Each input includes an informal target, its source context and assumptions, a formal target in Lean, and relevant formal dependencies.
The model links aligned regions on the informal and formal sides and then labels each link's faithfulness.
This measures whether the model locates the mathematical content behind each defect and judges it correctly.
By making individual correspondences explicit, align-then-label makes the model's judgments traceable and supports systematic analysis of errors.

We introduce \benchmark{}, a heterogeneous benchmark with 460 items and 4,982 annotated alignment links built from five mathematical formalization sources.
The sources vary in alignment density and structure, while the links range from multi-span correspondences to regions with no counterpart, with 25.5\% involving context or assumptions.
The annotations distinguish defects such as missing conditions from benign informal commentary that need not appear in the formal statement.
We also introduce \contextgap{}, a 146-item companion set that withholds informal context and assumptions.
Of these items, 85 become undecidable and 61 remain decidable, testing whether models recognize when a faithfulness judgment depends on missing information.

Evaluation of \numModels{} models reveals frequent failures to identify the complete set of defects.
GPT-6 Astra, the strongest model on defect F1, scores only 47.8\% and identifies every defect in only 34.3\% of defective items.
It fails to align 68.6\% of defective context links.
Models also tend to produce coarser alignments than humans, merging distinct mathematical objects and missing crucial correspondences.
When informal context and assumptions are withheld, the best model recognizes only 28.2\% of undecidable items: models often give definite verdicts that the available information cannot support.

We summarize our contributions as follows:
\textbf{(1)} We operationalize faithfulness evaluation as a context-aware \emph{align-then-label} problem, linking informal and formal contents across targets, context, and assumptions while allowing \emph{undecidable} judgments for underspecified inputs.
\textbf{(2)} We create \benchmark{}, 
a heterogeneous benchmark with fine-grained correspondence annotations for evaluating defect and underspecification identification.
\textbf{(3)} We empirically evaluate \numModels{} models, revealing recurring failures to identify all defects, align contextual dependencies, and recognize undecidable cases.

\section{Illustrative Overview}
\label{sec:illustrative-overview}

\input{figures/illustrative-example}

We illustrate the importance of faithful formalization with an example from ProofNet-Verified \citep{an2026ground}, shown in Figure~\ref{fig:motivating-example}. 
The natural language target proves a property over a vector space $V$ that is intended to hold over the real numbers $\mathbb{R}$ and the complex numbers $\mathbb{C}$.
However, an unfaithful formalization found in the dataset instantiates the scalar field over $\mathbb{C}$, while omitting $\mathbb{R}$. The faithful formalization, on the other hand, correctly introduces \texttt{\{F\} [RCLike F]}, which quantifies over both $\mathbb{R}$ and $\mathbb{C}$ and is thereby faithful. Though relatively small, at the theory scale this error compounds if later theorems refer to this one in the dependency graph. In that case, the error propagates to downstream items, rendering entire sections of the overall theory unusable.

The example also reveals why such judgments are difficult.
Taken in isolation, the natural language target is immensely under-specified: the symbols $V$ and $\mathcal{L}(V)$ are not defined in the natural language target. Further, even if we know that $V$ is a vector space, the target does not specify over which field it is defined. Such ambiguities cause both models and human annotators to output incorrect faithfulness labels. In this particular example, the unfaithful formalization is the ground truth released with ProofNet \citep{azerbayev2023proofnet}, which the expert annotators of ProofNet-Verified \citep{an2026ground} later judged unfaithful and corrected to the faithful version shown here. The fact that a formalization accepted in one expert-curated benchmark was flagged as defective in another shows how easily these ambiguities slip past human review. To address this ambiguity gap, careful annotation is required to identify relevant context outside the natural language target. Doing so foregrounds these gaps, making explicit \emph{how} ambiguities are resolved, rather than asserting that the target is unambiguous.

Accordingly, we pose faithfulness evaluation as being dependent on the necessary context $C$ rather than only the targets $T$, and further require links between natural language and formal phrases to localize how we arrived at the verdict. To illustrate the importance of context, Figure~\ref{fig:motivating-example} shows various natural language contexts ($C_N$) that appear earlier in the book: for example how $F$ denotes $\mathbb{R}$ or $\mathbb{C}$ ($C_N[1]$), and what the notation $\mathcal{L}(V)$ ($C_N[2]$) means. We further require explicit mappings between phrase-level segments to show \emph{precisely where} the formalization is faithful or not. For instance, the link from $C_N[1]$ to the formal target \texttt{[InnerProductSpace $\mathbb{C}$ V] [FiniteDimensional $\mathbb{C}$ V]} ($T_F[2]$) shows precisely that $T_F[2]$ is unfaithful (because it is missing $\mathbb{R}$), and that the link to \texttt{RCLike F} is faithful. The faithful link starting at $C_N[2]$ shows why each specific part of the formalization is faithful in both versions. These links in this example tell us that only one section of the unfaithful formalization is the issue, not the entire item, allowing this way of annotation to \emph{localize} issues rather than giving whole-item verdicts.

\begin{figure}[t]
    \includegraphics[width=\linewidth]{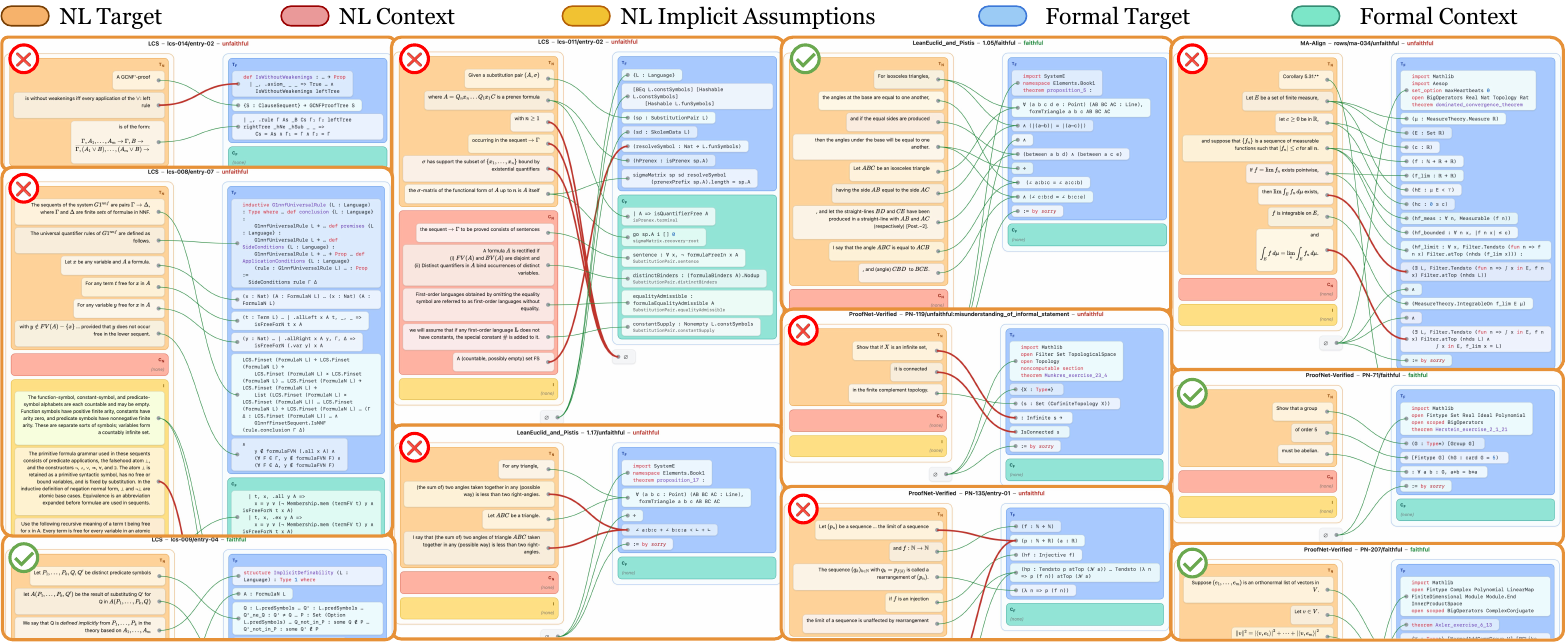}
    \caption{
        A bird's-eye-view illustration of eleven samples (with truncation) from \benchmark, where each segment from natural language (NL) or formal language is bucketed into Target, Context, or Implicit Assumptions.
        Segments from both sides are matched or unconnected (connected to $\varnothing$):
        faithful links are indicated by green lines, while unfaithful links are marked in red, implying that the overall Lean formalization is unfaithful to the NL source.
    }
    \label{fig:illustrative-overview-bev}
\end{figure}

While Figure~\ref{fig:motivating-example} shows the alignment of a single example, Figure~\ref{fig:illustrative-overview-bev} provides a bird's-eye view over samples drawn from ProofNet-Verified \citep{an2026ground}, LCS-Bench \citep{feng2026lcsbench}, MA-Align \citep{patel2026mathatlas}, LeanEuclid \citep{murphy2024euclid}, and Pistis \citep{mao2026pistis}.
Across these sources, \benchmark{} provides comprehensive and fine-grained annotation of alignments so that every unfaithful defect (illustrated with red lines) is explicitly captured and can be systematically analyzed.

\section{Problem Formulation}
\label{sec:problem-formulation}

In Section~\ref{sec:illustrative-overview}, we demonstrated that a faithfulness judgment depends on context beyond the target statement and is most useful when it locates each discrepancy.
Motivated by this observation, we formulate faithfulness evaluation as a novel align-then-label problem in which the evaluator aligns informal text with Lean code and then labels each resulting link using the context that determines the meaning of both statements. We first specify the inputs to this process, including the targets, their surrounding context, and any implicit assumptions. Figure~\ref{fig:notation} summarizes the notation.

\begin{figure}[t]
    \centering
    \captionsetup{skip=3pt}
    \includegraphics[width=\linewidth]{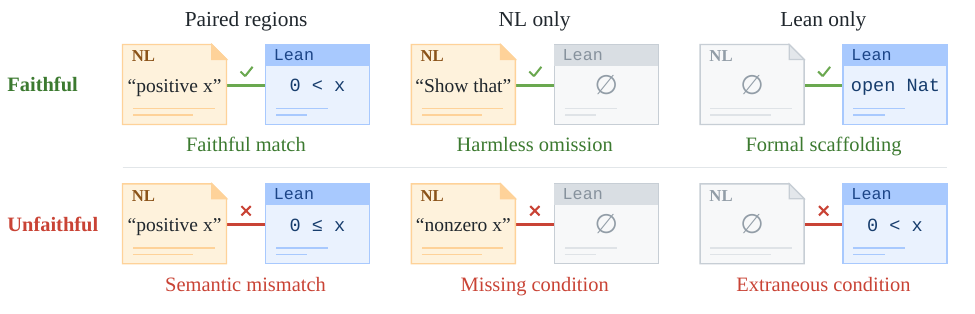}
    \caption{Schematic examples of the six link types.
    Rows give the label, and columns give which regions the link contains.
    $\varnothing$ marks an absent counterpart.}
    \label{fig:link-types}
\end{figure}

\begin{figure}[t]
    \centering
    \captionsetup{skip=6pt}
    {\footnotesize
    $\begin{array}{@{}l@{\quad}r@{\ \ }c@{\ \ }l@{\qquad}l@{\quad}r@{\ \ }c@{\ \ }l@{}}
        \text{(Item)} & x & ::= & (T_N, C_N, I, T_F, C_F) &
        \text{(NL region)} & \sigma & \subseteq & \text{one of } T_N, C_N, I \\
        \text{(NL target)} & T_N & ::= & \mathsf{statement} &
        \text{(Lean region)} & \tau & \subseteq & \text{one of } T_F, C_F \\
        \text{(NL context)} & C_N & ::= & [(\mathsf{quote}, \mathsf{source}), \ldots] &
        \text{(Link)} & a & ::= & (\sigma, \tau) \neq (\varnothing, \varnothing) \\
        \text{(Implicit assumptions)} & I & ::= & [\mathsf{assumption}, \ldots] &
        \text{(Label)} & \ell & ::= & 1 \mid 0 \mid \bot \\
        \text{(Lean target)} & T_F & ::= & \mathsf{Lean\ code} &
        \text{(Finding)} & y & ::= & (a, \ell) \\
        \text{(Lean context)} & C_F & ::= & [\mathsf{declaration}, \ldots] &
        \text{(Finding set)} & Y, \widehat{Y} & ::= & \{y_1, y_2, \ldots\} \\
        \text{(Matched pairs)} & \Pi & \subseteq & \widehat{Y} \times Y &
        \text{(Weight)} & w(\widehat{y}, y) & \in & [0, 1]
    \end{array}$
    \par}
    \caption{Notation for align-then-label.
    $Y$ denotes annotated findings and $\widehat{Y}$ predicted ones.}
    \label{fig:notation}
\end{figure}

\subsection{Inputs}
\label{sec:formulation-instances}

As Figure~\ref{fig:motivating-example} shows, the same statement can admit different interpretations when read in isolation.
Each item therefore pairs the informal and formal targets with the context that sufficiently disambiguates their respective meanings. We distinguish context available from the natural language source from context available from the formalization, and separately represent assumptions needed to resolve ambiguities that neither source makes explicit.

The informal target $T_N$ is the statement to be formalized, and the formal target $T_F$ is its Lean formalization \citep{moura2021lean4}.
The informal context $C_N$ quotes the definitions, notation, and standing assumptions that $T_N$ relies on.
The formal context $C_F$ lists the declarations that $T_F$ references, such as Mathlib definitions \citep{mathlib2020}.
Where the source leaves an interpretive choice open, such as an unstated convention, the implicit assumptions $I$ state it explicitly.

\subsection{Labeled Alignments}
\label{sec:formulation-alignments}

\noindent\textbf{Regions and links.}~~
To tie each judgment to the expressions responsible for it, an alignment is built from regions connected by links.
A \emph{region} is a set of selected characters, which may consist of several disjoint spans.
A \emph{link} pairs an informal region $\sigma$ with a formal region $\tau$.
Because missing and extraneous conditions have no counterpart by definition, either endpoint may be empty, but not both.
A one-sided link $(\sigma,\varnothing)$ records informal material with no formal counterpart, and $(\varnothing,\tau)$ records formal material with no informal counterpart.
Regions may overlap and participate in several links.
To evaluate links involving context and assumptions separately from those between the targets, we assign each link a role by its endpoints.
A link is \emph{implicit} if an endpoint lies in $I$, \emph{context} if it has no endpoint in $I$ but one in $C_N$ or $C_F$, and \emph{target} otherwise.

\noindent\textbf{Labels.}~~
Labels record the faithfulness judgment for each link, with a third label for links that the supplied material cannot settle.
Each link carries a faithfulness label $\ell$, faithful ($1$) or unfaithful ($0$).
For a paired link, the label records whether both regions express the same mathematical content, and for a one-sided link, whether the material may lack a counterpart, yielding the six link types in Figure~\ref{fig:link-types}.
We additionally consider a setting in which $C_N$ and $I$ are withheld, so that some links become \emph{undecidable} ($\bot$).
Such a link admits reasonable interpretations with different faithfulness judgments, and a definite label would commit to one that the supplied material does not establish.

\noindent\textbf{Findings and verdicts.}~~
Whole-item verdicts are derived from labeled links, so every verdict can be traced to the supporting links.
A labeled link is a \emph{finding} $y=(a,\ell)$, and align-then-label outputs a finding set.
This set contains faithful findings as well as defects.
A faithful verdict rests on them, and they show whether an unreported defect was misjudged or never aligned.
$Y$ denotes the reference finding set of an item and $\widehat{Y}$ denotes a predicted one.
The verdict is unfaithful if any finding is unfaithful, otherwise undecidable if any finding is undecidable, and faithful otherwise.
A definite defect therefore decides the item even when there are undecidable links.

\subsection{Evaluation Metrics}
\label{sec:formulation-metrics}

\noindent\textbf{F1 scores.}~~
F1 scores measure how much of the reference links are recovered with correct labels.
Credit is graded by endpoint overlap, so selecting a broad span that happens to enclose a reference region earns only partial credit.
Within each item, predicted links are matched one-to-one to reference links by overlap, ignoring labels, so a correctly located but mislabeled link is still matched.
Let $\Pi$ be the resulting set of matched pairs and $w(\widehat{y},y)$ the overlap weight of a pair $(\widehat{y},y)\in\Pi$.
For a label set $S\subseteq\{0,1,\bot\}$, let $Y_S$ and $\widehat{Y}_S$ be the reference and predicted findings whose labels lie in $S$.
The scores are then defined as follows:
\begin{equation*}
    \begin{array}{r@{\ }l@{\qquad}r@{\ }l}
        \textit{Credit}_S = & \textstyle\sum_{(\widehat{y},y)\in\Pi,\ \widehat{\ell}=\ell\in S} w(\widehat{y},y), &
        \textit{Prec}_S = & \textstyle\frac{\textit{Credit}_S}{|\widehat{Y}_S|}, \\[6pt]
        \textit{Rec}_S = & \textstyle\frac{\textit{Credit}_S}{|Y_S|}, &
        \textit{F1}_S = & \textstyle\frac{2\cdot\textit{Prec}_S\cdot\textit{Rec}_S}{\textit{Prec}_S+\textit{Rec}_S}.
    \end{array}
\end{equation*}
Credits and counts are summed over all items before the ratios are taken.
\emph{Joint F1} ($S=\{0,1,\bot\}$) measures recovery and correct labeling of all links.
Catching faithfulness errors matters most, so we separately calculate \emph{defect F1} ($S=\{0\}$), which isolates the recovery of defects.
\emph{Undecidable-link F1} ($S=\{\bot\}$) measures recognition of links that cannot be decided with the provided material.
We also report \emph{defect F1@0.5} which counts a defect only when it is located substantially.
It uses the same definitions with $S=\{0\}$, but a pair enters $\Pi$ only if the overlap has at least half recall and precision at every endpoint, and each pair contributes $1$ in place of $w(\widehat{y},y)$.

\noindent\textbf{Item-level rates.}~~
F1 pools links across items and therefore does not show whether any single item has been fully diagnosed.
The \emph{all-defects rate} is the fraction of items with at least one reference defect in which every reference defect is matched to a prediction labeled unfaithful.
\emph{Verdict accuracy} is the fraction of items whose verdict inferred from $\widehat{Y}$ matches the reference verdict.
To measure whether an evaluator correctly raises undecidable when the context is insufficient, the \emph{undecidable-item rate} restricts verdict accuracy to undecidable items.
The Appendix specifies the overlap weight, the matching, and content preprocessing.

\section{Benchmark}
\label{sec:benchmark}

\begin{figure}[t]
    \centering
    \setlength{\parskip}{0pt}
    \captionsetup{skip=3pt}
    \input{figures/dataset-composition}
    \caption{Composition and input lengths for LCS-Bench (LCS), ProofNet-Verified (PN-V), MA-Align (MA), and LeanEuclid--Pistis (LE-P).
    In (c), centers are group means; area $\propto$ count.}
    \label{fig:dataset-composition}
    \nextfloat
    \input{figures/endpoint-composition}
    \caption{Structure of annotated alignments.
    (a) counts faithful/unfaithful links per endpoint pair.}
    \label{fig:endpoint-composition}
\end{figure}

\subsection{Benchmark Construction}
\label{sec:benchmark-construction}

\noindent\textbf{Sources and target pairs.}~~
\benchmark{} draws on five sources, namely LeanEuclid \citep{murphy2024euclid}, Pistis \citep{mao2026pistis}, MA-Align \citep{patel2026mathatlas}, LCS-Bench \citep{feng2026lcsbench}, and ProofNet-Verified \citep{an2026ground}.
For each source, we use the provided informal--formal pairs as targets and manually construct their contexts from their textbook sources \citep[e.g.,][]{gallier2015logic}.
Since Pistis refines LeanEuclid, we report the two as one group, LeanEuclid--Pistis.

\noindent\textbf{Context selection and annotation.}~~
Four graduate-level experts in Lean and mathematics, all with prior Lean research experience, annotated the benchmark over about eight person-months using a browser-based visual tool in three steps.
First, they select broad regions on the informal and formal sides, connect corresponding regions or mark counterparts as absent, and assign labels.
Second, for unfaithful links, they narrow the selected regions to isolate the misalignment.
Each unfaithful region is made as precise as possible while remaining coherent.
Third, they supply context for each mathematical object required to understand $T_N$ and $T_F$.
On the informal side, this consists of complete source segments, such as paragraphs or book subsections, containing the relevant background.
Annotators also add assumptions to resolve ambiguity or underspecification left by the source context.
On the formal side, annotators gather the referenced declarations and other dependencies needed to understand the selected targets.
Context selection covers the mathematical dependencies of both targets.
A second annotator then reviews and refines every item, and the completed annotations retain 81.9\% alignment-link overlap with the initial versions, counting labels, under the evaluation scorer.

\noindent\textbf{Constructing \contextgap{}.}~~
\contextgap{} tests whether evaluators recognize when the supplied material is insufficient to settle faithfulness.
Each item keeps the two targets and the formal context of a \benchmark{} item but withholds $C_N$ and $I$. These items are annotated separately, with undecidable labels on links that depend on the withheld material.
The set contains undecidable items, whose targets admit readings with different faithfulness judgments without that material.
It also contains decidable controls that are unambiguous under common assumptions.
For each undecidable item, we also construct an opposite-verdict counterpart whose informal context reverses the verdict of the original item, forming a triplet of the undecidable item, its original item, and the counterpart.

\subsection{Dataset Statistics}
\label{sec:benchmark-statistics}

\benchmark{} contains 460 full-context items, with 164 faithful and 296 unfaithful whole-item verdicts.
Its 4,982 annotated links, 4,232 faithful and 750 unfaithful, range from 2 to 34 per item, and 25.5\% of them involve context or assumptions.
Figure~\ref{fig:dataset-composition} summarizes source composition and input lengths for LCS-Bench (LCS), ProofNet-Verified (PN-V), MA-Align (MA), and LeanEuclid--Pistis (LE-P), and Figure~\ref{fig:endpoint-composition} summarizes endpoint types and links per item.
\contextgap{} is counted separately and contains 146 items, of which 85 are undecidable, 12 faithful, and 49 unfaithful.
Its 1,481 links are 954 faithful, 103 unfaithful, and 424 undecidable ones, and the Appendix gives its composition.

\subsection{Evaluation Tasks and Protocol}
\label{sec:benchmark-protocol}

We evaluate models on two align-then-label tasks, which differ in whether the context that fixes the interpretation is supplied.
In both tasks, the model returns a labeled alignment rather than a verdict, and the item verdict is inferred from its findings by the rule in Section~\ref{sec:formulation-alignments}.
Both tasks are scored with the metrics defined in Section~\ref{sec:formulation-metrics}, and the \contextgap{} triplets connect them by pairing each undecidable item with full-context versions of both readings.

\noindent\textbf{Full-context alignment.}~~
This task measures recovery of both faithful links and defects under a fixed interpretation.
On the 460 \benchmark{} items, the model receives all five input components of Section~\ref{sec:formulation-instances} and labels each link faithful or unfaithful.
Each nonempty endpoint is a set of character spans from the targets, from the context, or from a single assumption.
The alignment also requires unmatched material.
An acceptable omission or addition therefore forms a faithful one-sided link, and a required condition missing from the formalization, including one stated only in the context, forms an unfaithful link with an empty formal endpoint.
Context is selected only where it supplies the material being assessed or fixes the interpretation of a link.
The model is instructed to align coherent mathematical contributions, such as the setup, hypotheses, and conclusion, rather than individual symbols, and to split a contribution that contains separable faithful and defective parts.
We report defect F1, together with defect F1@0.5, joint F1, and the all-defects rate.

\noindent\textbf{Context-omitted alignment.}~~
This task measures whether the model recognizes when the supplied material cannot settle faithfulness.
On the 146 \contextgap{} items, the model receives only $T_N$, $T_F$, and $C_F$ and may additionally label a link undecidable.
An undecidable link connects the visible informal phrase whose interpretation is unresolved with the formal material whose faithfulness depends on it.
Undecidable items and decidable controls use the same prompt, which does not reveal which kind an item is, so the model must itself determine whether the supplied material settles each link.
We report undecidable-link F1, the undecidable-item rate, and verdict accuracy over all 146 items, split into undecidable items and decidable controls.
On the 85 triplets, comparing each undecidable item with its two full-context versions separates recognition of underspecification from correct judgment once the missing context is supplied; we therefore report verdict accuracy for both the context-omitted and full-context versions.

\section{Experiments}
\label{sec:experiments}

\begin{figure}[t]
    \centering
    \captionsetup{skip=3pt}
    \includegraphics[width=\linewidth]{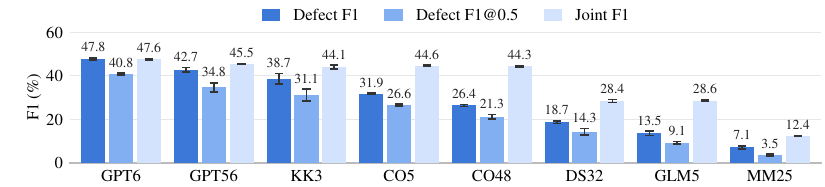}
    \caption{Full-context alignment F1 per model.}
    \label{fig:full-alignment}
    \nextfloat
    \includegraphics[width=\linewidth]{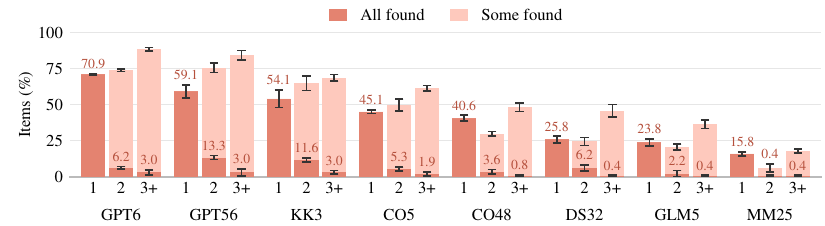}
    \caption{Defect identification by the number of defects in an item, over 133, 75, and 88 items with 1, 2, and 3+ defects.
    Full bar height counts items with at least one defect found.}
    \label{fig:defects-found}
\end{figure}

\subsection{Experimental Setup}
\label{sec:experimental-setup}
We evaluate \numModels{} models on the tasks in Section~\ref{sec:benchmark-protocol}, with a 32,768-token output limit.
The models are GPT-6 Astra (GPT6), GPT-5.6 Sol (GPT56), Kimi K3 (KK3), Opus 5 (CO5), Opus 4.8 (CO48), DeepSeek V3.2 (DS32), GLM-5 (GLM5), and MiniMax M2.5 (MM25).
The evaluation covers all 460 \benchmark{} items and all 146 \contextgap{} items, with three repetitions of full-context and context-omitted alignment per model.
Figures report means with sample standard deviations.
Alignment tasks provide text-search and answer-validation tools, with no mathematical correctness feedback.
Scoring details are provided in the Appendix.

\subsection{Research Questions and Findings}
\label{sec:research-questions}

\begin{figure}[t]
    \centering
    \captionsetup{skip=3pt,position=bottom}
    \begin{minipage}[t]{.65\linewidth}
        \centering\vspace{0pt}
        \input{figures/model-comparisons}
    \end{minipage}\hfill
    \begin{minipage}[t]{.33\linewidth}
        \centering\vspace{0pt}
        \includegraphics[width=\linewidth]{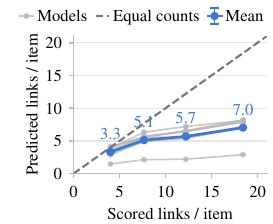}
    \end{minipage}
    \par\nointerlineskip\noindent
    \begin{minipage}[t]{.65\linewidth}
        \caption{Defect outcomes and causes of missed links for GPT6, GPT56, KK3, and CO5, in percent.
        In (b), a disconnected link has both endpoints selected in separate predictions, and an overmerged link is covered by a prediction already matched to another link.}
        \label{fig:model-comparisons}
    \end{minipage}\hfill
    \begin{minipage}[t]{.33\linewidth}
        \nextfloat
        \caption{Predicted versus annotated links per item, with items grouped by annotated link count.}
        \label{fig:alignment-link-scaling}
    \end{minipage}
\end{figure}

\begin{figure}[t]
    \centering
    \captionsetup{skip=3pt}
    \input{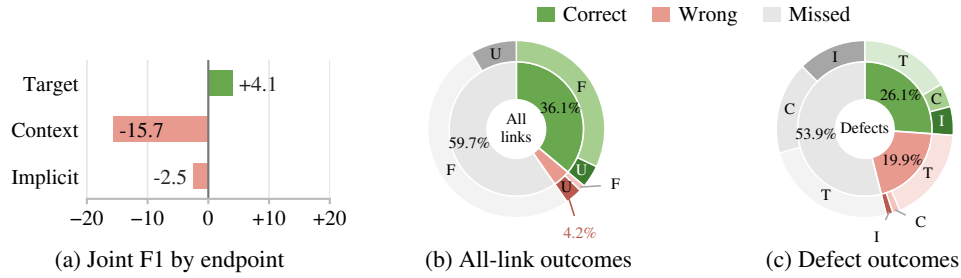}
    \caption{Alignment scores and outcomes across all models.
    (a) Joint F1 relative to the pooled 37.6\%, in percentage points.
    Outer rings split outcomes by label in (b), faithful (F) or unfaithful (U), and by role in (c), target (T), context (C), or implicit (I).}
    \label{fig:endpoint-scores}
\end{figure}

\noindent\textbf{RQ1: Can models identify faithfulness defects?}~~
Models rarely find all the defects in an item, and most of the defects they miss are never aligned at all.
Figure~\ref{fig:full-alignment} shows that in full-context alignment, even the strongest model, GPT-6 Astra, reaches a defect F1 of only 47.8\%, and its score drops further under the stricter F1@0.5.
Whole-item verdicts would hide this gap.
Astra's inferred verdicts reach 93.1\% balanced accuracy, averaging its accuracy on faithful and unfaithful items, although it identifies every defect in only 34.3\% of defective items.
As Figure~\ref{fig:defects-found} shows, this rate falls to 4.5\% on items with two or more defects.
On two-defect items, Astra finds exactly one defect in 68\% of cases, against 48\% under independent detection at the same per-defect rate, so it finds both far less often than independence predicts.
Figure~\ref{fig:defect-analysis} attributes most failures to defects that are never aligned rather than to defects aligned but wrongly labeled, so the bottleneck lies mainly in \textit{locating} defects rather than in \textit{judging} them.
These omissions persist across attempts, since on average nearly two thirds of defects are never identified in any of the three repetitions.

\noindent\textbf{RQ2: How well do models identify and assess contextual links?}~~
Models often fail to connect the targets to the context they depend on, so most contextual defects go unassessed.
Figure~\ref{fig:endpoint-relative} shows that links involving context have the lowest joint F1 of all link roles, 15.7 percentage points below the overall score.
The deficit comes mostly from missing links, since roughly three quarters of contextual defects are never aligned, and defects involving implicit assumptions are omitted in much the same way.
Supplying the surrounding material therefore does not ensure that a model uses it, which is a central failure for theory-scale assessment.

\begin{figure}[t]
    \centering
    \captionsetup{skip=3pt,position=bottom}
    \begin{minipage}[t]{.50\linewidth}
        \vspace{0pt}
        \centering
        \includegraphics[width=\linewidth]{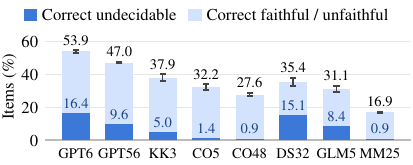}
        \caption{Context-omitted verdict accuracy on \contextgap{} by verdict type, as percentages of all 146 items, so bar height gives overall accuracy.}
        \label{fig:contextgap-correct}
    \end{minipage}\hfill
    \begin{minipage}[t]{.48\linewidth}
        \vspace{0pt}
        \centering
        \captionsetup{type=table,position=top}
        \caption{GPT6 verdicts for the 85 triplets, under full-context alignment of the faithful (Full F) and unfaithful (Full U) versions and under context-omitted alignment (Gap).
        Columns show predicted verdicts, with ND for undecidable.}
        \label{tab:contextgap-judgments}
        \small
        \setlength{\tabcolsep}{2.5pt}
        \renewcommand{\arraystretch}{1.2}
        \begin{tabular*}{\linewidth}[b]{@{}@{\extracolsep{\fill}}lrrr@{}}
            \toprule
            Input & F & U & ND \\
            \midrule
            \input data/contextgap_judgments.tex
            \bottomrule
        \end{tabular*}
    \end{minipage}
\end{figure}

\noindent\textbf{RQ3: Can models recognize underspecification when relevant context is absent?}~~
Models seldom recognize that missing context leaves a judgment open, and instead commit to one reading.
Figure~\ref{fig:contextgap-correct} shows that in context-omitted alignment, most correct verdicts come from the decidable controls, while every model struggles to identify the undecidable items.
A definite verdict on an undecidable item could stem from two different failures, misjudging the item or not noticing that its judgment depends on the withheld context.
The paired comparison separates the two by stating each reading explicitly.
In full-context alignment, where the supplied context fixes the interpretation, Astra judges both the faithful and the unfaithful version correctly in 96.9\% of triplets, as Table~\ref{tab:contextgap-judgments} shows, so it can judge each reading once that reading is stated.
In context-omitted alignment, however, it usually supplies an unstated assumption of its own, and for 52 of the 85 triplets it judges both versions correctly yet gives a definite verdict in every repetition.
The failure therefore lies in recognizing that a judgment depends on context the model has not been given.

\noindent\textbf{RQ4: Can models produce fine-grained mathematical alignments?}~~
Current models cannot yet produce alignments at the granularity of the annotations, and most annotated links receive no matching prediction.
As Figure~\ref{fig:alignment-link-scaling} shows, in full-context alignment the number of predicted links grows much more slowly than the number of annotated links, and roughly 60\% of annotated links remain unaligned on average.
Figure~\ref{fig:unmatched-links} separates two kinds of failure behind these gaps.
About a third on average are \emph{overmerged links}, in which a single prediction spans several annotated links and collapses their separate mathematical relationships into one label.
In the rest, models omit one or both endpoints, or select both endpoints in separate links without connecting them.

\section{Related Work}
\label{sec:related-work}

\noindent\textbf{Autoformalization.}~~
Large language models have enabled substantial progress in
autoformalization~\citep{wu2022autoformalization,weng2025survey}, with benchmarks spanning competition, undergraduate, and
library-scale Lean
formalization~\citep{zheng2022minif2f,azerbayev2023proofnet,ying2024leanworkbook,gao2025herald,yu2025formalmath}.
Recent work has extended this setting to theory-scale, dependency-aware
formalization~\citep{min2026theory,feng2026lcsbench,poiroux2025reliable,rammal2026scale,liu2025rethinking,wang2026aria}.
Formalization also serves as an intermediate step, guiding provers with formal proof
sketches~\citep{jiang2023dsp} or filtering informal solutions whose formalizations are
inconsistent~\citep{zhou2024dontrust}, and retrieval of related formalizations helps keep
library-scale autoformalization consistent~\citep{zhang2024consistent}.

Pipelines that synthesize training data for provers filter autoformalized statements with
model-based quality scores, checks for contradictory hypotheses, and whole-statement
faithfulness judgments~\citep{xin2024deepseekprover,lin2025goedelprover}.
Such filters decide whether to keep a statement but not which of its parts is unfaithful.
Audits of existing benchmarks and formalizations have identified
unfaithful statements, including missing or added hypotheses, that
compilation alone does not
reveal~\citep{poiroux2025reliable,an2026ground,meek2026numerical},
highlighting the need for evaluation in context.

\noindent\textbf{Faithfulness evaluation.}~~
Existing evaluations largely judge candidates at the statement level.
Reference-based methods compare against gold formalizations using
equivalence or structural
similarity~\citep{liu2025rethinking,poiroux2025reliable,murphy2024euclid,liu2025gted,liu2026assess};
source-based methods test consistency with the informal
statement~\citep{li2024consistency,han2026shadowbench,mohammad2026gap}; and
model-based evaluators directly assess informal--formal
alignment~\citep{lu2025formalalign,peng2026criticlean,wang2026aria,yu2025mathesis,zhang2026beyond},
with related work on evaluator reliability and error
localization~\citep{chen2026reform,patel2026mathatlas,wang2026formalrx}.
Beyond statements, recent work asks whether formal proofs follow their
source arguments~\citep{mao2026pistis} and whether autoformalizers silently
repair invalid reasoning steps~\citep{cornish2026faithformbench}. Our benchmark instead prioritizes fine-grained judgment, making labeled informal--formal links the prediction
target and including missing content and undecidable items.

\noindent\textbf{LLM judges and error localization.}~~
LLM judges can agree with human preferences on whole responses while showing systematic
biases~\citep{zheng2023judging}.
Work on mathematical reasoning moves supervision and evaluation from final answers to
individual steps, training reward models on step-level feedback and asking models to
identify the earliest erroneous step~\citep{lightman2024verify,zheng2025processbench}.
Fine-grained hallucination benchmarks label erroneous spans in generated text, including
spans that retrieved context does not support~\citep{mishra2024fava,niu2024ragtruth}.
These settings localize errors within a single text, whereas align-then-label attaches
each judgment to a pair of informal and formal regions and to the context that fixes
their meaning.

\noindent\textbf{Fine-grained evaluation and alignment.}~~
We draw on fine-grained NLP evaluation that aligns
chunks and marks unaligned material, localizes errors hidden by
aggregate scores, and separates ambiguity from missing
information~\citep{agirre2016ists,pagnoni2021frank,min2023factscore,bhaskar2023ambiqt,thorne2018fever}.
Word aligners link tokens across languages~\citep{jalilisabet2020simalign}, and
monolingual aligners extend this to phrases of varying length~\citep{lan2021monolingual}.
Align-then-label links natural language to Lean instead, where a region may consist of
several disjoint spans and each link carries a faithfulness label.

\noindent\textbf{Ambiguity and abstention.}~~
Question answering benchmarks ask models to abstain on unanswerable
questions~\citep{rajpurkar2018squad2} or to recover every plausible reading of an
ambiguous one~\citep{min2020ambigqa}, and abstention in LLMs more broadly has been
surveyed~\citep{wen2025abstention}.
Language models are often well calibrated about whether they know an
answer~\citep{kadavath2022know}, yet often fail to recognize ambiguity in
entailment~\citep{liu2023ambient}, and code generation systems ask clarifying questions
when requirements are ambiguous~\citep{mu2024clarifygpt}.
\contextgap{} poses the corresponding question for faithfulness: the evaluator must recognize
that the supplied material leaves the interpretation of a link open, a property of the
input rather than of the model's own knowledge.

\section{Conclusion}
\label{sec:conclusion}

We formulated theory-scale faithfulness evaluation as a context-aware align-then-label problem and introduced \benchmark{} and \contextgap{} to evaluate defect identification and recognition of underspecification.
Models frequently miss defects, overlook contextual links, and give definite judgments when essential information is absent.
These findings highlight detailed alignment and reasoning about contextual dependencies as central challenges for reliable faithfulness assessment.
Future work should develop evaluators that systematically trace dependencies, search for additional defects, and identify the missing definitions or assumptions needed to resolve undecidable judgments.

\subsection*{AI Use Statement}

For the MA-Align items, Codex and Claude Code proposed candidate spans, and a human annotator refined every proposed span.
No other annotation step used generative AI, and a second annotator reviewed and refined every item, as the Appendix describes.

We also used generative AI tools, including Codex, to assist with refining evaluation methodology, implementing evaluation and analysis code, and interpreting experimental results.
We also used these tools for literature search, manuscript drafting and revision, and figure preparation.
Apart from the MA-Align span proposals, we did not use generative AI to generate data, including the authored opposite-verdict counterparts, to develop the align-then-label formulation, to propose hypotheses, to clean or reformat the dataset, or to translate text.
Formulating and proving mathematical claims and qualitative data analysis do not apply to this work.
AI-assisted drafts and figures were revised through iterative author review, and reported numerical summaries were checked against saved experimental results.
The authors take responsibility for the final content, claims, and artifacts of this work.

\subsection*{Ethics Statement}

\benchmark{} and \contextgap{} use mathematical texts and formalizations from the cited sources.
Researchers with access to the cited sources can request the full context text, exactly as annotated, from the authors.
The paper quotes selected excerpts to explain the annotated judgments.

\subsection*{Reproducibility Statement}

The textbook context text is available on request, as the Ethics Statement describes.

Section~\ref{sec:benchmark-construction} describes the sources, context construction, and annotation.
The evaluation tasks are specified in Section~\ref{sec:benchmark-protocol}; settings and prompts appear in Appendices~\ref{app:model-settings} and~\ref{app:prompts}.
Appendix~\ref{app:dataset-details} provides detailed dataset statistics, and Appendix~\ref{app:metric-details} defines endpoint matching, label credit, and aggregation across repetitions.
Appendix~\ref{app:additional-results} reports repeated-attempt results and additional analyses.

\subsection*{Acknowledgments}

This research was supported by an Amazon Research Award (Fall 2025), the 2026 Amazon Nova AI Challenge: Trusted Software Agents, and the U.S. Department of Energy under the Genesis Mission, Award No. DE-SC0026865.
The views and conclusions contained herein are those of the authors and should not be interpreted as necessarily representing the official policies of the sponsors.

\bibliographystyle{iclr2027_conference}
\bibliography{references}

\begin{thebibliography}{59}
\providecommand{\natexlab}[1]{#1}
\providecommand{\url}[1]{\texttt{#1}}
\expandafter\ifx\csname urlstyle\endcsname\relax
  \providecommand{\doi}[1]{doi: #1}\else
  \providecommand{\doi}{doi: \begingroup \urlstyle{rm}\Url}\fi

\bibitem[Agirre et~al.(2016)Agirre, Gonzalez-Agirre, Lopez-Gazpio, Maritxalar, Rigau, and Uria]{agirre2016ists}
Eneko Agirre, Aitor Gonzalez-Agirre, I{\~n}igo Lopez-Gazpio, Montse Maritxalar, German Rigau, and Larraitz Uria.
\newblock {SemEval}-2016 task 2: Interpretable semantic textual similarity.
\newblock In \emph{Proceedings of the 10th International Workshop on Semantic Evaluation (SemEval-2016)}, pp.\  512--524. Association for Computational Linguistics, 2016.

\bibitem[An et~al.(2026)An, Min, Si, and Bastani]{an2026ground}
One An, Marcus~J. Min, Xujie Si, and Osbert Bastani.
\newblock Ground false: Dissecting errors in formal mathematics benchmarks.
\newblock In \emph{3rd AI for Math Workshop: Toward Self-Evolving Scientific Agents at ICML 2026}, 2026.

\bibitem[Axler(2015)]{axler2015linear}
Sheldon Axler.
\newblock \emph{Linear Algebra Done Right}.
\newblock Undergraduate Texts in Mathematics. Springer, 3rd edition, 2015.

\bibitem[Azerbayev et~al.(2023)Azerbayev, Piotrowski, Schoelkopf, Ayers, Radev, and Avigad]{azerbayev2023proofnet}
Zhangir Azerbayev, Bartosz Piotrowski, Hailey Schoelkopf, Edward~W. Ayers, Dragomir Radev, and Jeremy Avigad.
\newblock {ProofNet}: Autoformalizing and formally proving undergraduate-level mathematics.
\newblock \emph{arXiv preprint arXiv:2302.12433}, 2023.

\bibitem[Bhaskar et~al.(2023)Bhaskar, Tomar, Sathe, and Sarawagi]{bhaskar2023ambiqt}
Adithya Bhaskar, Tushar Tomar, Ashutosh Sathe, and Sunita Sarawagi.
\newblock Benchmarking and improving text-to-{SQL} generation under ambiguity.
\newblock In \emph{Proceedings of the 2023 Conference on Empirical Methods in Natural Language Processing}, pp.\  7053--7074. Association for Computational Linguistics, 2023.

\bibitem[Chen et~al.(2026)Chen, Wu, Chen, Zhao, Song, Li, Fan, Liu, and Liao]{chen2026reform}
Guoxin Chen, Jing Wu, Xinjie Chen, Wayne~Xin Zhao, Ruihua Song, Chengxi Li, Kai Fan, Dayiheng Liu, and Minpeng Liao.
\newblock {ReForm}: Reflective autoformalization with prospective bounded sequence optimization.
\newblock In \emph{International Conference on Learning Representations}, 2026.

\bibitem[Cornish et~al.(2026)Cornish, Ghinassi, Yeh, Liu, Xu, Yin, Wagner, Li, Teh, and Ong]{cornish2026faithformbench}
Rob Cornish, Iacopo Ghinassi, Po-Hung Yeh, Shuqi Liu, Qiyuan Xu, Haoxuan Yin, Dominik Wagner, Wenda Li, Yee~Whye Teh, and Luke Ong.
\newblock {FaithformBench}: Benchmarking faithfulness of mathematical chain-of-thought autoformalisation.
\newblock \emph{arXiv preprint arXiv:2608.10916}, 2026.

\bibitem[de~Moura \& Ullrich(2021)de~Moura and Ullrich]{moura2021lean4}
Leonardo de~Moura and Sebastian Ullrich.
\newblock The {Lean 4} theorem prover and programming language.
\newblock In \emph{Automated Deduction -- {CADE} 28}, Lecture Notes in Computer Science, pp.\  625--635. Springer, 2021.

\bibitem[Feng et~al.(2026)Feng, Pu, An, Bastani, Zhang, Huang, Si, and Li]{feng2026lcsbench}
Yuming Feng, Frederick Pu, One An, Osbert Bastani, Li~Zhang, Jiani Huang, Xujie Si, and Ziyang Li.
\newblock Theory-scale auto-formalization of logics for computer science.
\newblock \emph{arXiv preprint arXiv:2606.26525}, 2026.

\bibitem[Fitzpatrick(2008)]{fitzpatrick2008euclid}
Richard Fitzpatrick.
\newblock {Euclid's Elements of Geometry}, 2008.
\newblock Greek text of J.~L. Heiberg with an English translation by Richard Fitzpatrick, revised and corrected edition, ISBN 978-0-6151-7984-1.

\bibitem[Gallier(2015)]{gallier2015logic}
Jean Gallier.
\newblock \emph{Logic for Computer Science: Foundations of Automatic Theorem Proving}.
\newblock Dover Publications, 2nd edition, 2015.

\bibitem[Gao et~al.(2025)Gao, Wang, Jiang, Gao, Qin, Xu, and Dong]{gao2025herald}
Guoxiong Gao, Yutong Wang, Jiedong Jiang, Qi~Gao, Zihan Qin, Tianyi Xu, and Bin Dong.
\newblock {Herald}: A natural language annotated {Lean 4} dataset.
\newblock In \emph{International Conference on Learning Representations}, 2025.

\bibitem[Han et~al.(2026)Han, Kim, Park, Cheon, Park, Jeon, Choe, Kong, Hur, Hwang, and Hyeon]{han2026shadowbench}
Hojae Han, Jongyoon Kim, Sanghyeok Park, Dongwook Cheon, Yeachan Park, Myeong~Jae Jeon, Sunjong Choe, Soonho Kong, Wonseok Hur, Seung-won Hwang, and Donghoon Hyeon.
\newblock {SHADOWBENCH}: Toward reliable automatic evaluation of semantic alignment in autoformalization.
\newblock \emph{arXiv preprint arXiv:2608.29270}, 2026.

\bibitem[Jalili~Sabet et~al.(2020)Jalili~Sabet, Dufter, Yvon, and Sch{\"u}tze]{jalilisabet2020simalign}
Masoud Jalili~Sabet, Philipp Dufter, Fran{\c{c}}ois Yvon, and Hinrich Sch{\"u}tze.
\newblock {SimAlign}: High quality word alignments without parallel training data using static and contextualized embeddings.
\newblock In \emph{Findings of the Association for Computational Linguistics: EMNLP 2020}, pp.\  1627--1643. Association for Computational Linguistics, 2020.

\bibitem[Jiang et~al.(2023)Jiang, Welleck, Zhou, Li, Liu, Jamnik, Lacroix, Wu, and Lample]{jiang2023dsp}
Albert~Q. Jiang, Sean Welleck, Jin~Peng Zhou, Wenda Li, Jiacheng Liu, Mateja Jamnik, Timoth{\'e}e Lacroix, Yuhuai Wu, and Guillaume Lample.
\newblock Draft, sketch, and prove: Guiding formal theorem provers with informal proofs.
\newblock In \emph{International Conference on Learning Representations}, 2023.

\bibitem[Kadavath et~al.(2022)Kadavath, Conerly, Askell, Henighan, Drain, Perez, Schiefer, Hatfield-Dodds, DasSarma, Tran-Johnson, Johnston, El-Showk, Jones, Elhage, Hume, Chen, Bai, Bowman, Fort, Ganguli, Hernandez, Jacobson, Kernion, Kravec, Lovitt, Ndousse, Olsson, Ringer, Amodei, Brown, Clark, Joseph, Mann, McCandlish, Olah, and Kaplan]{kadavath2022know}
Saurav Kadavath, Tom Conerly, Amanda Askell, Tom Henighan, Dawn Drain, Ethan Perez, Nicholas Schiefer, Zac Hatfield-Dodds, Nova DasSarma, Eli Tran-Johnson, Scott Johnston, Sheer El-Showk, Andy Jones, Nelson Elhage, Tristan Hume, Anna Chen, Yuntao Bai, Sam Bowman, Stanislav Fort, Deep Ganguli, Danny Hernandez, Josh Jacobson, Jackson Kernion, Shauna Kravec, Liane Lovitt, Kamal Ndousse, Catherine Olsson, Sam Ringer, Dario Amodei, Tom Brown, Jack Clark, Nicholas Joseph, Ben Mann, Sam McCandlish, Chris Olah, and Jared Kaplan.
\newblock Language models (mostly) know what they know.
\newblock \emph{arXiv preprint arXiv:2207.05221}, 2022.

\bibitem[Lan et~al.(2021)Lan, Jiang, and Xu]{lan2021monolingual}
Wuwei Lan, Chao Jiang, and Wei Xu.
\newblock Neural semi-{Markov} {CRF} for monolingual word alignment.
\newblock In \emph{Proceedings of the 59th Annual Meeting of the Association for Computational Linguistics and the 11th International Joint Conference on Natural Language Processing (Volume 1: Long Papers)}, pp.\  6815--6828. Association for Computational Linguistics, 2021.

\bibitem[Lang(2002)]{lang2002algebra}
Serge Lang.
\newblock \emph{Algebra}.
\newblock Graduate Texts in Mathematics. Springer, 2002.

\bibitem[Li et~al.(2024)Li, Wu, Li, Wei, Yang, Zhang, and Ma]{li2024consistency}
Zenan Li, Yifan Wu, Zhaoyu Li, Xinming Wei, Fan Yang, Xian Zhang, and Xiaoxing Ma.
\newblock Autoformalize mathematical statements by symbolic equivalence and semantic consistency.
\newblock In \emph{Advances in Neural Information Processing Systems}, volume~37, pp.\  53598--53625. Curran Associates, Inc., 2024.

\bibitem[Lightman et~al.(2024)Lightman, Kosaraju, Burda, Edwards, Baker, Lee, Leike, Schulman, Sutskever, and Cobbe]{lightman2024verify}
Hunter Lightman, Vineet Kosaraju, Yura Burda, Harri Edwards, Bowen Baker, Teddy Lee, Jan Leike, John Schulman, Ilya Sutskever, and Karl Cobbe.
\newblock Let's verify step by step.
\newblock In \emph{International Conference on Learning Representations}, 2024.

\bibitem[Lin et~al.(2025)Lin, Tang, Lyu, Wu, Lin, Yang, Li, Xia, Chen, Arora, and Jin]{lin2025goedelprover}
Yong Lin, Shange Tang, Bohan Lyu, Jiayun Wu, Hongzhou Lin, Kaiyu Yang, Jia Li, Mengzhou Xia, Danqi Chen, Sanjeev Arora, and Chi Jin.
\newblock {Goedel-Prover}: A frontier model for open-source automated theorem proving.
\newblock \emph{arXiv preprint arXiv:2502.07640}, 2025.

\bibitem[Liu et~al.(2023)Liu, Wu, Michael, Suhr, West, Koller, Swayamdipta, Smith, and Choi]{liu2023ambient}
Alisa Liu, Zhaofeng Wu, Julian Michael, Alane Suhr, Peter West, Alexander Koller, Swabha Swayamdipta, Noah Smith, and Yejin Choi.
\newblock We're afraid language models aren't modeling ambiguity.
\newblock In \emph{Proceedings of the 2023 Conference on Empirical Methods in Natural Language Processing}, pp.\  790--807. Association for Computational Linguistics, 2023.

\bibitem[Liu et~al.(2025{\natexlab{a}})Liu, Zheng, Lu, Cao, and Yan]{liu2025rethinking}
Qi~Liu, Xinhao Zheng, Xudong Lu, Qinxiang Cao, and Junchi Yan.
\newblock Rethinking and improving autoformalization: Towards a faithful metric and a dependency retrieval-based approach.
\newblock In \emph{International Conference on Learning Representations}, 2025{\natexlab{a}}.

\bibitem[Liu et~al.(2026)Liu, Zhu, Dong, Liu, Guo, Liu, Chen, and Luo]{liu2026assess}
Xiaoyang Liu, Tao Zhu, Zineng Dong, Yuntian Liu, Qingfeng Guo, Zhaoxuan Liu, Yu~Chen, and Tao Luo.
\newblock {ASSESS}: A semantic and structural evaluation framework for statement similarity.
\newblock In \emph{International Conference on Learning Representations}, 2026.

\bibitem[Liu et~al.(2025{\natexlab{b}})Liu, Zhu, Liu, Chen, Liu, Guo, Zhang, Bao, and Luo]{liu2025gted}
Yuntian Liu, Tao Zhu, Xiaoyang Liu, Yu~Chen, Zhaoxuan Liu, Qingfeng Guo, Jiashuo Zhang, Kangjie Bao, and Tao Luo.
\newblock Generalized tree edit distance ({GTED}): A faithful evaluation metric for statement autoformalization.
\newblock \emph{arXiv preprint arXiv:2507.07399}, 2025{\natexlab{b}}.

\bibitem[Lu et~al.(2025)Lu, Wan, Huang, Xiong, Liu, and Guo]{lu2025formalalign}
Jianqiao Lu, Yingjia Wan, Yinya Huang, Jing Xiong, Zhengying Liu, and Zhijiang Guo.
\newblock {FormalAlign}: Automated alignment evaluation for autoformalization.
\newblock In \emph{International Conference on Learning Representations}, 2025.

\bibitem[Mao et~al.(2026)Mao, Zhong, Arekar, Feng, An, Huang, Si, and Li]{mao2026pistis}
Tadd Mao, Tianjun Zhong, Dhruva Arekar, Yuming Feng, One An, Jiani Huang, Xujie Si, and Ziyang Li.
\newblock Does the proof prove it that way? {F}aithful formalization of {Elements} proofs.
\newblock \emph{arXiv preprint arXiv:2608.15432}, 2026.

\bibitem[Meek et~al.(2026)Meek, Ge, Xiang, Chess, and Ilin]{meek2026numerical}
Theodore Meek, Siyuan Ge, Di~Qiu Xiang, Simon Chess, and Vasily Ilin.
\newblock Formalizing numerical analysis: An agent pipeline and quality audit beyond kernel acceptance.
\newblock \emph{arXiv preprint arXiv:2606.14000}, 2026.

\bibitem[Min et~al.(2026)Min, He, Li, Yi, Malik, Gupta, Si, and Bastani]{min2026theory}
Marcus~J. Min, Mike He, Zhaoyu Li, Zixuan Yi, Sharad Malik, Aarti Gupta, Xujie Si, and Osbert Bastani.
\newblock Theory-level autoformalization: From isolated statements to unified formal knowledge bases.
\newblock \emph{arXiv preprint arXiv:2607.13292}, 2026.
\newblock ICML 2026 position paper, published as ``Position: The Case for Theory-Level Autoformalization''.

\bibitem[Min et~al.(2020)Min, Michael, Hajishirzi, and Zettlemoyer]{min2020ambigqa}
Sewon Min, Julian Michael, Hannaneh Hajishirzi, and Luke Zettlemoyer.
\newblock {AmbigQA}: Answering ambiguous open-domain questions.
\newblock In \emph{Proceedings of the 2020 Conference on Empirical Methods in Natural Language Processing}, pp.\  5783--5797. Association for Computational Linguistics, 2020.

\bibitem[Min et~al.(2023)Min, Krishna, Lyu, Lewis, Yih, Koh, Iyyer, Zettlemoyer, and Hajishirzi]{min2023factscore}
Sewon Min, Kalpesh Krishna, Xinxi Lyu, Mike Lewis, Wen-tau Yih, Pang~Wei Koh, Mohit Iyyer, Luke Zettlemoyer, and Hannaneh Hajishirzi.
\newblock {FActScore}: Fine-grained atomic evaluation of factual precision in long form text generation.
\newblock In \emph{Proceedings of the 2023 Conference on Empirical Methods in Natural Language Processing}, pp.\  12076--12100. Association for Computational Linguistics, 2023.

\bibitem[Mishra et~al.(2024)Mishra, Asai, Balachandran, Wang, Neubig, Tsvetkov, and Hajishirzi]{mishra2024fava}
Abhika Mishra, Akari Asai, Vidhisha Balachandran, Yizhong Wang, Graham Neubig, Yulia Tsvetkov, and Hannaneh Hajishirzi.
\newblock Fine-grained hallucination detection and editing for language models.
\newblock In \emph{Conference on Language Modeling}, 2024.

\bibitem[Mohammad \& Sheikh(2026)Mohammad and Sheikh]{mohammad2026gap}
Noor Islam~S. Mohammad and Tamim Sheikh.
\newblock The faithfulness gap: Certifying semantic equivalence between natural-language and formal mathematical statements.
\newblock \emph{arXiv preprint arXiv:2606.16541}, 2026.

\bibitem[Mu et~al.(2024)Mu, Shi, Wang, Yu, Zhang, Wang, Liu, and Wang]{mu2024clarifygpt}
Fangwen Mu, Lin Shi, Song Wang, Zhuohao Yu, Binquan Zhang, ChenXue Wang, Shichao Liu, and Qing Wang.
\newblock {ClarifyGPT}: A framework for enhancing {LLM}-based code generation via requirements clarification.
\newblock \emph{Proceedings of the ACM on Software Engineering}, 1\penalty0 (FSE):\penalty0 2332--2354, 2024.
\newblock \doi{10.1145/3660810}.

\bibitem[Murphy et~al.(2024)Murphy, Yang, Sun, Li, Anandkumar, and Si]{murphy2024euclid}
Logan Murphy, Kaiyu Yang, Jialiang Sun, Zhaoyu Li, Anima Anandkumar, and Xujie Si.
\newblock Autoformalizing {Euclidean} geometry.
\newblock In \emph{Proceedings of the 41st International Conference on Machine Learning}, volume 235 of \emph{Proceedings of Machine Learning Research}, pp.\  36847--36893. PMLR, 2024.

\bibitem[Niu et~al.(2024)Niu, Wu, Zhu, Xu, Shum, Zhong, Song, and Zhang]{niu2024ragtruth}
Cheng Niu, Yuanhao Wu, Juno Zhu, Siliang Xu, KaShun Shum, Randy Zhong, Juntong Song, and Tong Zhang.
\newblock {RAGTruth}: A hallucination corpus for developing trustworthy retrieval-augmented language models.
\newblock In \emph{Proceedings of the 62nd Annual Meeting of the Association for Computational Linguistics (Volume 1: Long Papers)}, pp.\  10862--10878. Association for Computational Linguistics, 2024.

\bibitem[Pagnoni et~al.(2021)Pagnoni, Balachandran, and Tsvetkov]{pagnoni2021frank}
Artidoro Pagnoni, Vidhisha Balachandran, and Yulia Tsvetkov.
\newblock Understanding factuality in abstractive summarization with {FRANK}: A benchmark for factuality metrics.
\newblock In \emph{Proceedings of the 2021 Conference of the North American Chapter of the Association for Computational Linguistics: Human Language Technologies}, pp.\  4812--4829. Association for Computational Linguistics, 2021.

\bibitem[Patel et~al.(2026)Patel, Arias, Babayan, Cochran, Libman, Mahmood, McCarty, Munoz, Willey, and Flanigan]{patel2026mathatlas}
Nilay Patel, Noah Arias, Davit Babayan, Victoria Cochran, Timothy Libman, Hafsah Mahmood, Liam McCarty, Soli Munoz, Laurel Willey, and Jeffrey Flanigan.
\newblock {MathAtlas}: A benchmark for autoformalization in the wild.
\newblock \emph{arXiv preprint arXiv:2605.14061}, 2026.

\bibitem[Peng et~al.(2026)Peng, Yao, Ma, Guo, Li, Zhang, Zhang, Zhang, Yu, Li, Liu, Xia, Shen, Wu, Cao, Zhang, Huang, Liu, and Zhang]{peng2026criticlean}
Zhongyuan Peng, Yifan Yao, Kaijing Ma, Shuyue Guo, Yizhe Li, Yichi Zhang, Chenchen Zhang, Yifan Zhang, Zhouliang Yu, Luming Li, Minghao Liu, Yihang Xia, Jiawei Shen, Yuchen Wu, Yixin Cao, Zhaoxiang Zhang, Wenhao Huang, Jiaheng Liu, and Ge~Zhang.
\newblock {CriticLean}: Critic-guided reinforcement learning for mathematical formalization.
\newblock In \emph{Proceedings of the 64th Annual Meeting of the Association for Computational Linguistics (Volume 1: Long Papers)}, pp.\  3049--3088. Association for Computational Linguistics, 2026.

\bibitem[Poiroux et~al.(2025)Poiroux, Weiss, Kun{\v{c}}ak, and Bosselut]{poiroux2025reliable}
Auguste Poiroux, Gail Weiss, Viktor Kun{\v{c}}ak, and Antoine Bosselut.
\newblock Reliable evaluation and benchmarks for statement autoformalization.
\newblock In \emph{Proceedings of the 2025 Conference on Empirical Methods in Natural Language Processing}, pp.\  17947--17969. Association for Computational Linguistics, 2025.

\bibitem[Rajpurkar et~al.(2018)Rajpurkar, Jia, and Liang]{rajpurkar2018squad2}
Pranav Rajpurkar, Robin Jia, and Percy Liang.
\newblock Know what you don't know: Unanswerable questions for {SQuAD}.
\newblock In \emph{Proceedings of the 56th Annual Meeting of the Association for Computational Linguistics (Volume 2: Short Papers)}, pp.\  784--789. Association for Computational Linguistics, 2018.

\bibitem[Rammal et~al.(2026)Rammal, Patel, Gloeckle, Hayat, Kempe, Munos, Arnal, and Cabannes]{rammal2026scale}
Ahmad Rammal, Niket Patel, Fabian Gloeckle, Amaury Hayat, Julia Kempe, Remi Munos, Charles Arnal, and Vivien Cabannes.
\newblock Formalizing mathematics at scale.
\newblock \emph{arXiv preprint arXiv:2605.29955}, 2026.

\bibitem[{The mathlib Community}(2020)]{mathlib2020}
{The mathlib Community}.
\newblock The {Lean} mathematical library.
\newblock In \emph{Proceedings of the 9th ACM SIGPLAN International Conference on Certified Programs and Proofs}, pp.\  367--381. Association for Computing Machinery, 2020.

\bibitem[Thorne et~al.(2018)Thorne, Vlachos, Christodoulopoulos, and Mittal]{thorne2018fever}
James Thorne, Andreas Vlachos, Christos Christodoulopoulos, and Arpit Mittal.
\newblock {FEVER}: A large-scale dataset for fact extraction and {VERification}.
\newblock In \emph{Proceedings of the 2018 Conference of the North American Chapter of the Association for Computational Linguistics: Human Language Technologies, Volume 1 (Long Papers)}, pp.\  809--819. Association for Computational Linguistics, 2018.

\bibitem[Wang et~al.(2026{\natexlab{a}})Wang, Xie, Wang, Gao, Yu, and Dong]{wang2026aria}
Hanyu Wang, Ruohan Xie, Yutong Wang, Guoxiong Gao, Xintao Yu, and Bin Dong.
\newblock {Aria}: An agent for retrieval and iterative auto-formalization via dependency graph.
\newblock In \emph{International Conference on Learning Representations}, 2026{\natexlab{a}}.

\bibitem[Wang et~al.(2026{\natexlab{b}})Wang, Huang, Wan, Zhu, Liu, Huang, and Guo]{wang2026formalrx}
Haocheng Wang, Baiyu Huang, Yingjia Wan, Xiao Zhu, Xiaoyang Liu, Yinya Huang, and Zhijiang Guo.
\newblock {FormalRx}: Rectify and {eXamine} semantic failures in autoformalization.
\newblock In \emph{Proceedings of the 43rd International Conference on Machine Learning}, 2026{\natexlab{b}}.

\bibitem[Wen et~al.(2025)Wen, Yao, Feng, Xu, Tsvetkov, Howe, and Wang]{wen2025abstention}
Bingbing Wen, Jihan Yao, Shangbin Feng, Chenjun Xu, Yulia Tsvetkov, Bill Howe, and Lucy~Lu Wang.
\newblock Know your limits: A survey of abstention in large language models.
\newblock \emph{Transactions of the Association for Computational Linguistics}, 13:\penalty0 529--556, 2025.

\bibitem[Weng et~al.(2025)Weng, Du, Li, Lu, Sun, Liu, and Zhang]{weng2025survey}
Ke~Weng, Lun Du, Sirui Li, Wangyue Lu, Haozhe Sun, Hengyu Liu, and Tiancheng Zhang.
\newblock Autoformalization in the era of large language models: A survey.
\newblock \emph{arXiv preprint arXiv:2505.23486}, 2025.

\bibitem[Wu et~al.(2022)Wu, Jiang, Li, Rabe, Staats, Jamnik, and Szegedy]{wu2022autoformalization}
Yuhuai Wu, Albert~Qiaochu Jiang, Wenda Li, Markus Rabe, Charles Staats, Mateja Jamnik, and Christian Szegedy.
\newblock Autoformalization with large language models.
\newblock In \emph{Advances in Neural Information Processing Systems}, volume~35, pp.\  32353--32368. Curran Associates, Inc., 2022.

\bibitem[Xin et~al.(2024)Xin, Guo, Shao, Ren, Zhu, Liu, Ruan, Li, and Liang]{xin2024deepseekprover}
Huajian Xin, Daya Guo, Zhihong Shao, Zhizhou Ren, Qihao Zhu, Bo~Liu, Chong Ruan, Wenda Li, and Xiaodan Liang.
\newblock {DeepSeek-Prover}: Advancing theorem proving in {LLMs} through large-scale synthetic data.
\newblock \emph{arXiv preprint arXiv:2405.14333}, 2024.

\bibitem[Ying et~al.(2024)Ying, Wu, Geng, Wang, Lin, and Chen]{ying2024leanworkbook}
Huaiyuan Ying, Zijian Wu, Yihan Geng, Jiayu Wang, Dahua Lin, and Kai Chen.
\newblock {Lean Workbook}: A large-scale {Lean} problem set formalized from natural language math problems.
\newblock In \emph{Advances in Neural Information Processing Systems}, volume~37, pp.\  105848--105863. Curran Associates, Inc., 2024.

\bibitem[Yu et~al.(2026)Yu, Zhong, Feng, Zhai, Mohit, Ng, Liu, Shou, Xiong, Zhou, Ong, Sugiarto, Zhang, Tai, Cao, Lu, Sun, Xu, Shen, and Li]{yu2025mathesis}
Xuejun Yu, Jianyuan Zhong, Zijin Feng, Pengyi Zhai, Roozbeh Mohit, Wei~Chong Ng, Haoxiong Liu, Ziyi Shou, Jing Xiong, Yudong Zhou, Claudia~Beth Ong, Austen~Jeremy Sugiarto, Yaoxi Zhang, Wai~Ming Tai, Huan Cao, Dongcai Lu, Jiacheng Sun, Qiang Xu, Xin Shen, and Zhenguo Li.
\newblock {Mathesis}: Towards formal theorem proving from natural languages.
\newblock In \emph{International Conference on Learning Representations}, 2026.

\bibitem[Yu et~al.(2025)Yu, Peng, Ding, Li, Peng, Liu, Zhang, Yuan, Xin, Huang, Wen, Zhang, and Liu]{yu2025formalmath}
Zhouliang Yu, Ruotian Peng, Keyi Ding, Yizhe Li, Zhongyuan Peng, Minghao Liu, Yifan Zhang, Zheng Yuan, Huajian Xin, Wenhao Huang, Yandong Wen, Ge~Zhang, and Weiyang Liu.
\newblock {FormalMATH}: Benchmarking formal mathematical reasoning of large language models.
\newblock \emph{arXiv preprint arXiv:2505.02735}, 2025.

\bibitem[Zhang et~al.(2026)Zhang, Gallardo~Candela, Murthy, Xie, Wang, and Raissi]{zhang2026beyond}
Ke~Zhang, Patricio Gallardo~Candela, Sudhir Murthy, Yi~Xie, Zhi Wang, and Maziar Raissi.
\newblock Beyond compilation: Evaluating faithful natural-language-to-{Lean} statement formalization.
\newblock \emph{arXiv preprint arXiv:2606.31002}, 2026.

\bibitem[Zhang et~al.(2024)Zhang, Quan, and Freitas]{zhang2024consistent}
Lan Zhang, Xin Quan, and Andre Freitas.
\newblock Consistent autoformalization for constructing mathematical libraries.
\newblock In \emph{Proceedings of the 2024 Conference on Empirical Methods in Natural Language Processing}, pp.\  4020--4033. Association for Computational Linguistics, 2024.

\bibitem[Zheng et~al.(2025)Zheng, Zhang, Zhang, Lin, Lu, Yu, Liu, Zhou, and Lin]{zheng2025processbench}
Chujie Zheng, Zhenru Zhang, Beichen Zhang, Runji Lin, Keming Lu, Bowen Yu, Dayiheng Liu, Jingren Zhou, and Junyang Lin.
\newblock {ProcessBench}: Identifying process errors in mathematical reasoning.
\newblock In \emph{Proceedings of the 63rd Annual Meeting of the Association for Computational Linguistics (Volume 1: Long Papers)}, pp.\  1009--1024. Association for Computational Linguistics, 2025.

\bibitem[Zheng et~al.(2022)Zheng, Han, and Polu]{zheng2022minif2f}
Kunhao Zheng, Jesse~Michael Han, and Stanislas Polu.
\newblock {MiniF2F}: A cross-system benchmark for formal {Olympiad}-level mathematics.
\newblock In \emph{International Conference on Learning Representations}, 2022.

\bibitem[Zheng et~al.(2023)Zheng, Chiang, Sheng, Zhuang, Wu, Zhuang, Lin, Li, Li, Xing, Zhang, Gonzalez, and Stoica]{zheng2023judging}
Lianmin Zheng, Wei-Lin Chiang, Ying Sheng, Siyuan Zhuang, Zhanghao Wu, Yonghao Zhuang, Zi~Lin, Zhuohan Li, Dacheng Li, Eric~P. Xing, Hao Zhang, Joseph~E. Gonzalez, and Ion Stoica.
\newblock Judging {LLM}-as-a-judge with {MT-Bench} and {Chatbot Arena}.
\newblock In \emph{Advances in Neural Information Processing Systems}, volume~36, pp.\  46595--46623. Curran Associates, Inc., 2023.

\bibitem[Zhou et~al.(2024)Zhou, Staats, Li, Szegedy, Weinberger, and Wu]{zhou2024dontrust}
Jin~Peng Zhou, Charles Staats, Wenda Li, Christian Szegedy, Kilian~Q. Weinberger, and Yuhuai Wu.
\newblock Don't trust: Verify -- grounding {LLM} quantitative reasoning with autoformalization.
\newblock In \emph{International Conference on Learning Representations}, 2024.

\end{thebibliography}

\clearpage
\appendix
\raggedbottom
\makeatletter
\setlength{\@fptop}{0pt}
\setlength{\@fpsep}{18pt}
\setlength{\@fpbot}{0pt plus 1fil}
\makeatother

\section{Dataset Details}
\label{app:dataset-details}

\subsection{Sources and Alignment Structure}
The 460 full-context items contain 4,982 annotated links (Table~\ref{tab:benchmark-sources}).
Although most items contain a defect, most individual links are faithful: a local discrepancy can make an otherwise aligned statement unfaithful.

\begin{table}[htbp]
    \centering
    \setlength{\tabcolsep}{3pt}
    \renewcommand{\arraystretch}{1.12}
    \caption{Full-context items and alignment links by source.}
    \label{tab:benchmark-sources}
    \small
    \begin{tabular*}{\textwidth}{@{}@{\extracolsep{\fill}}lrrrrrr@{}}
        \toprule
        & \multicolumn{3}{c}{Items} & \multicolumn{3}{c}{Links} \\
        \cmidrule(lr){2-4}\cmidrule(lr){5-7}
        Source & Total & Faithful & Unfaithful & Total & Faithful & Unfaithful \\
        \midrule
        LCS-Bench & 196 & 47 & 149 & 2,218 & 1,883 & 335 \\
        ProofNet-Verified & 77 & 31 & 46 & 436 & 374 & 62 \\
        MA-Align & 122 & 42 & 80 & 1,470 & 1,142 & 328 \\
        LeanEuclid--Pistis & 65 & 44 & 21 & 858 & 833 & 25 \\
        \midrule
        Total & 460 & 164 & 296 & 4,982 & 4,232 & 750 \\
        \bottomrule
    \end{tabular*}
\end{table}

The sources also differ in alignment structure (Table~\ref{tab:benchmark-properties}).
ProofNet-Verified items have fewer links, while one-sided links are most common in MA-Align and LeanEuclid--Pistis.
LCS-Bench contains most of the multi-span links, which have more than one text span on at least one endpoint.

\begin{table}[htbp]
    \centering
    \setlength{\tabcolsep}{3pt}
    \renewcommand{\arraystretch}{1.12}
    \caption{Alignment density and structure by source.}
    \label{tab:benchmark-properties}
    \small
    \begin{tabular*}{\textwidth}{@{}@{\extracolsep{\fill}}lrrrrrr@{}}
        \toprule
        & \multicolumn{4}{c}{Links per item} & & \\
        \cmidrule(lr){2-5}
        Source & Min. & Median & Mean & Max. & One-sided & Multi-span \\
        \midrule
        LCS-Bench & 2 & 11 & 11.32 & 23 & 128 & 1,130 \\
        ProofNet-Verified & 2 & 5 & 5.66 & 12 & 32 & 37 \\
        MA-Align & 2 & 10.5 & 12.05 & 34 & 359 & 139 \\
        LeanEuclid--Pistis & 5 & 13 & 13.20 & 28 & 257 & 192 \\
        \midrule
        Total & 2 & 10 & 10.83 & 34 & 776 & 1,498 \\
        \bottomrule
    \end{tabular*}
\end{table}

\subsection{Input Lengths}
Table~\ref{tab:input-lengths} reports the median and range of character counts by input component.
Counts use Unicode codepoints, include whitespace, and sum over the supplied documents assigned to each component.
An absent component has length zero.
\begin{table}[htbp]
    \centering
    \setlength{\tabcolsep}{3pt}
    \renewcommand{\arraystretch}{1.12}
    \caption{Input lengths in characters: median above [minimum, maximum].}
    \label{tab:input-lengths}
    \small
    \begin{tabular*}{\textwidth}{@{}@{\extracolsep{\fill}}lrrrr@{}}
        \toprule
        Component & LCS-Bench & ProofNet-Verified & MA-Align & LeanEuclid--Pistis \\
        \midrule
        Informal target $T_N$ & \shortstack[r]{392\\{[59, 1,874]}} & \shortstack[r]{114\\{[46, 377]}} & \shortstack[r]{195.5\\{[25, 1,661]}} & \shortstack[r]{404\\{[64, 1,037]}} \\[3pt]
        Formal target $T_F$ & \shortstack[r]{854\\{[152, 3,166]}} & \shortstack[r]{244\\{[104, 488]}} & \shortstack[r]{299\\{[96, 1,181]}} & \shortstack[r]{275\\{[124, 719]}} \\[3pt]
        Source context $C_N$ & \shortstack[r]{13,850\\{[246, 42,316]}} & \shortstack[r]{98\\{[0, 495]}} & \shortstack[r]{163\\{[0, 1,829]}} & \shortstack[r]{245\\{[0, 1,064]}} \\[3pt]
        Formal context $C_F$ & \shortstack[r]{6,656.5\\{[498, 18,641]}} & \shortstack[r]{3,411\\{[15, 15,800]}} & \shortstack[r]{1,002\\{[60, 9,138]}} & \shortstack[r]{1,893\\{[896, 6,691]}} \\[3pt]
        Assumptions $I$ & \shortstack[r]{0\\{[0, 2,623]}} & \shortstack[r]{0\\{[0, 614]}} & \shortstack[r]{0\\{[0, 844]}} & \shortstack[r]{0\\{[0, 161]}} \\[3pt]
        \bottomrule
    \end{tabular*}
\end{table}

Figure~\ref{fig:dataset-input-length} groups nearby lengths by complete-linkage clustering with Chebyshev distance in log space, with a maximum within-group length ratio of 1.5 on each axis.
Bubble centers are the groups' mean target and context lengths, and areas are proportional to item counts.

\subsection{\contextgap{}}
Table~\ref{tab:omission-composition} gives the composition of the 146 \contextgap{} items:
85 undecidable items paired with their original full-context counterparts, 12 faithful controls, and 49 unfaithful controls.
Their annotations contain 1,481 links: 954 faithful, 103 unfaithful, and 424 undecidable.
Undecidable links occur in all 85 undecidable items and in 33 unfaithful controls.
An item can therefore contain both an established defect and a link that cannot be judged from the available context.
The original full-context items of the 85 undecidable items comprise 72 faithful and 13 unfaithful items.
Their opposite-verdict counterparts are constructed, and their informal context or assumptions are an authored interpretation rather than a quotation from the source.
The 54 counterparts that are meaningful as standalone items are also \benchmark{} items, and the other 31 are additional evaluation inputs counted in neither set.
Each counterpart passed an independent review of context support and link quality.

\begin{table}[htbp]
    \centering
    \setlength{\tabcolsep}{3pt}
    \renewcommand{\arraystretch}{1.12}
    \caption{\contextgap{} items and link labels by source (F: faithful; U: unfaithful; ND: undecidable).}
    \label{tab:omission-composition}
    \small
    \begin{tabular*}{\textwidth}{@{}@{\extracolsep{\fill}}lrrrrrr@{}}
        \toprule
        & \multicolumn{3}{c}{Item verdicts} & \multicolumn{3}{c}{Link labels} \\
        \cmidrule(lr){2-4}\cmidrule(lr){5-7}
        Source & F & U & ND & F & U & ND \\
        \midrule
        LCS-Bench & 2 & 7 & 39 & 341 & 19 & 274 \\
        ProofNet-Verified & 7 & 10 & 13 & 95 & 15 & 20 \\
        MA-Align & 3 & 32 & 23 & 375 & 69 & 115 \\
        LeanEuclid--Pistis & 0 & 0 & 10 & 143 & 0 & 15 \\
        \midrule
        Total & 12 & 49 & 85 & 954 & 103 & 424 \\
        \bottomrule
    \end{tabular*}
\end{table}

\subsection{Annotation Process}
\label{app:annotation-process}
Each item passes through two expert annotators.
The four annotators are graduate-level experts in Lean and mathematics with prior experience in Lean research, and the annotation took about eight person-months.
One annotator constructs an item through the three steps in Section~\ref{sec:benchmark-construction}, and a second annotator then reviews and refines it.
Under the scorer used for model evaluation, the completed annotations retain 93.5\% span overlap and 89.5\% alignment-link overlap with the initial versions, so review refines the annotations rather than rewriting them.
This review covers every \benchmark{} item, every \contextgap{} item, and every opposite-verdict counterpart.

Generative AI assisted only with span proposals for the MA-Align items.
Codex and Claude Code proposed candidate spans for these items, and an annotator refined every proposed span.
No other annotation step used generative AI.

The AI-proposed spans show no sign of favoring the evaluated GPT and Claude models.
Codex and Claude Code run GPT and Claude models, and four of the eight evaluated models are GPT or Claude models.
On MA-Align, the only source with AI-proposed spans, these four models reach on average 70\% of their joint F1 on the other three sources, the same proportion as the other four models (Table~\ref{tab:source-results}).
Their defect F1 proportions also fall within the range of the other four models.

\section{Evaluation Details}
\label{app:metric-details}

\subsection{Model Providers and Generation Settings}
\label{app:model-settings}
All eight models are accessed through Amazon Bedrock Runtime in the \texttt{us-east-1} region.
Table~\ref{tab:model-settings} gives the exact model identifiers and API interfaces.

\begin{table}[htbp]
    \centering
    \small
    \setlength{\tabcolsep}{3pt}
    \renewcommand{\arraystretch}{1.12}
    \caption{Model identifiers and request settings on Amazon Bedrock. Medium denotes an explicitly requested reasoning effort; default denotes the provider's reasoning configuration. Both Opus models use adaptive thinking.}
    \label{tab:model-settings}
    \begin{tabular*}{\textwidth}{@{}@{\extracolsep{\fill}}llll@{}}
        \toprule
        Model & Bedrock model identifier & API & Reasoning \\
        \midrule
        GPT-6 Astra & \texttt{global.openai.gpt-6-astra} & Responses v1 & Medium \\
        GPT-5.6 Sol & \texttt{global.openai.gpt-5.6-sol} & Responses v1 & Medium \\
        Kimi K3 & \texttt{global.moonshotai.kimi-k3} & Responses v1 & Medium \\
        Opus 5 & \texttt{global.anthropic.claude-opus-5} & ConverseStream & Medium \\
        Opus 4.8 & \texttt{global.anthropic.claude-opus-4-8} & ConverseStream & Medium \\
        DeepSeek V3.2 & \texttt{deepseek.v3.2} & ConverseStream & Default \\
        GLM-5 & \texttt{zai.glm-5} & ConverseStream & Default \\
        MiniMax M2.5 & \texttt{minimax.minimax-m2.5} & ConverseStream & Default \\
        \bottomrule
    \end{tabular*}
\end{table}

\noindent\textbf{Generation parameters.}
Every model request has a 32,768-token output limit.
The two GPT models and Kimi K3 use \texttt{reasoning.effort=medium}; the two Opus models use adaptive thinking with \texttt{output\_config.effort=medium}.
Temperature and top-$p$ are left at provider defaults; requests specify neither a random seed nor custom stop sequences.
Final JSON is requested through the prompt rather than constrained JSON-schema decoding.

\noindent\textbf{Repetitions.}
Full-context and context-omitted alignment each have three repetitions per model, as do Astra's opposite-verdict full-context judgments.
Each selected-region direction has one response per model and item.

\noindent\textbf{Tools and prompts.}
Models can call \texttt{find\_text} to locate exact strings in supplied documents and \texttt{validate\_answer} to check JSON structure, labels, and span coordinates.
These tools provide no mathematical correctness feedback.
Appendix~\ref{app:prompts} gives the system prompts and user-message format.

\subsection{Content and Link Scores}
\noindent\textbf{Scored mathematical content.}
We trim import and scope commands, theorem/lemma title prefixes, and \texttt{sorry} proof placeholders from predicted and reference formal endpoints.
Mixed selections retain their mathematical spans and labels.
If trimming empties a formal endpoint, a remaining informal endpoint forms a one-sided link; a finding with neither endpoint remaining is removed.
In the selected-region task, formal regions consisting entirely of these spans and their incident predictions are omitted from scoring.
Whole-item verdicts inferred from alignments use the retained findings.
This preprocessing leaves 4,768 full-context reference links and 1,443 \contextgap{} reference links.

\noindent\textbf{Endpoint overlap and link weight.}
Regions are sets of selected character positions.
Overlaps are computed within each document and never bridge gaps between disjoint spans or different documents.
For a predicted region $P$ and an annotated region $G$, the Dice overlap is $D(P,G)=2|P\cap G|/(|P|+|G|)$, and two empty regions have overlap one.
A predicted link $\widehat{a}=(\widehat{\sigma},\widehat{\tau})$ and an annotated link $a=(\sigma,\tau)$ form an eligible pair only if they leave the same endpoints empty and overlap at every nonempty endpoint.
The weight of an eligible pair is the geometric mean of its two endpoint overlaps,
\begin{equation*}
    w(\widehat{y},y)=\sqrt{D(\widehat{\sigma},\sigma)\,D(\widehat{\tau},\tau)},
\end{equation*}
so the weight of a one-sided pair is the square root of the overlap at its nonempty endpoint.
Ineligible pairs have weight zero.
Selected-region tasks require the supplied region identifier and geometry to agree, then score only the counterpart overlap.

\noindent\textbf{One-to-one matching.}
Within each item, the matched pairs $\Pi$ form a maximum-weight one-to-one matching between predicted and annotated links.
Each link is matched at most once, only eligible pairs may be matched, and the total weight of the matched pairs is maximized.
We solve this linear assignment problem exactly with SciPy's \texttt{linear\_sum\_assignment} and discard any assigned pair that is not eligible.
Labels enter neither the weights, the eligibility test, nor tie-breaking, so every label-specific score and link-category breakdown reuses the same $\Pi$.
Unmatched links earn no credit.

\noindent\textbf{Credit and aggregation.}
For a label set $S$, $\textit{Credit}_S$ sums $w(\widehat{y},y)$ over the matched pairs whose two labels agree and lie in $S$.
Precision and recall therefore share the same credit, and $\textit{F1}_S=2\,\textit{Credit}_S/(|\widehat{Y}_S|+|Y_S|)$.
Joint F1 uses $S=\{0,1,\bot\}$, defect F1 uses $S=\{0\}$, and undecidable-link F1 uses $S=\{\bot\}$.
A prediction labeled faithful that is matched to an annotated defect thus lowers defect recall without entering defect precision.
Within each repetition, credits and counts are summed over all items before the ratios are taken, and reported values are means and sample standard deviations over the three repetitions.
$\textit{F1}_S$ is zero if exactly one of the pooled populations $\widehat{Y}_S$ and $Y_S$ is empty and undefined if both are empty.
Figure~\ref{fig:endpoint-scores} sums credits and counts before computing joint F1 for each link category and the overall baseline.
The three categories partition the links: an assumption endpoint determines the implicit category; otherwise a context endpoint determines the context category, and the remaining links belong to the target category.

The alignment outcomes in Figure~\ref{fig:endpoint-scores} count annotated links under the same assignment: correctly labeled matches, incorrectly labeled matches, and unmatched references.
In Figure~\ref{fig:endpoint-outcomes}, each outcome is subdivided by the reference link's faithful or unfaithful label; Figure~\ref{fig:defect-role-outcomes} restricts the population to unfaithful links and subdivides each outcome by link role.
These are discrete match outcomes; partially overlapping endpoints can receive a match.

\noindent\textbf{Thresholded defect F1.}
Defect F1@0.5 changes eligibility and credit but keeps the populations of defect F1.
A pair is eligible only if both links leave the same endpoints empty and, at every nonempty endpoint, $|P\cap G|\geq 0.5\,|P|$ and $|P\cap G|\geq 0.5\,|G|$.
The label-blind one-to-one matching first maximizes the number of matched pairs and, among matchings of that size, their total weight $w$.
We implement this order by adding to every eligible weight a constant larger than the number of links on the smaller side, which exceeds any possible difference in total weight.
Each matched pair with both links labeled unfaithful earns one unit of credit, and the precision and recall denominators are $|\widehat{Y}_{\{0\}}|$ and $|Y_{\{0\}}|$.

\subsection{Complete Sets and Repeated Attempts}
\noindent\textbf{Identification of every defect.}
The All defects score uses the 296 full-context items containing at least one annotated unfaithful link.
An item succeeds when every unfaithful link has a matched prediction labeled unfaithful under the standard label-blind assignment.
Each match requires positive overlap at every nonempty endpoint and agreement on empty endpoints.
The score measures complete defect recall at the item level; additional predictions are assessed by defect F1.
Figure~\ref{fig:defects-found} reports this rate separately for items with one, two, and at least three defects, using each group's item count as its denominator.

\noindent\textbf{Verdict accuracy and undecidable-item rate.}
Verdicts are inferred from the retained predicted findings: unfaithful if any finding is labeled unfaithful, otherwise undecidable if any finding is labeled undecidable, and faithful otherwise.
Verdict accuracy compares this verdict with the annotated verdict over a given item set.
Balanced verdict accuracy averages verdict accuracy over faithful and unfaithful items, and for GPT-6 Astra under full-context alignment it is 93.1\% on average over the three repetitions.
The undecidable-item rate uses the 85 undecidable \contextgap{} items, so an item succeeds only when no finding is labeled unfaithful and at least one is labeled undecidable.

\noindent\textbf{Best-of-three F1 and complete-item pass@3.}
For each item, best-of-three F1 selects the response with the highest item-level F1 separately for defect and joint scoring, then applies the same count-and-credit aggregation as the main results.
Selection uses the annotations. On defect-free items, a response predicting no defects is preferred.
Complete-item pass@3 records whether at least one of the three responses correctly identifies the entire required set.
The defect criterion is the All defects criterion above; the all-links criterion requires every annotated link to have a matched prediction with the correct label.
Both use the primary assignment and its endpoint-overlap requirements.

\noindent\textbf{Exact counterpart sets.}
For selected regions, exact counterpart-set identification requires equality of the predicted and annotated region sets, including document identities, selected character locations, and empty counterparts. Labels and ordering do not affect this diagnostic.

\subsection{Context Judgments}
\noindent\textbf{Paired context judgments.}
Each of the 85 undecidable \contextgap{} items is paired with the original full-context item used to construct it.
The \contextgap{} item and its parent share $T_N$, $T_F$, and $C_F$; the parent also supplies $C_N$ and $I$.
Their full-context parents comprise 72 faithful and 13 unfaithful items.
Table~\ref{tab:contextgap-paired-results} compares correct verdicts for these same 85 items under each condition.

For Astra, we also evaluate a full-context interpretation with the opposite faithfulness verdict for every undecidable item.
Both full-context versions preserve $T_N$, $T_F$, and $C_F$ and supply $C_N$ and $I$ to settle their respective interpretations.
Table~\ref{tab:contextgap-judgments} reports the mean and sample standard deviation across three repetitions for all three context conditions.
An item counts toward the 52 persistent failures when both full-context interpretations are judged correctly in all three repetitions and all three context-omitted alignment responses give definite verdicts.

\subsection{Alignment Diagnostics}
\noindent\textbf{Number of links.}
Figure~\ref{fig:alignment-link-scaling} groups full-context items by their annotated link counts: 2--5, 6--10, 11--15, and at least 16.
The alignment-size curves use all 460 items and compare predicted and annotated link counts after content trimming.

\noindent\textbf{Coverage and labeling.}
The diagnostics use the same content trimming, trials, and label-blind one-to-one assignment as the main scores.
Counts and weights are pooled over items as in the main scores.
Let $|Y|$ and $|\widehat{Y}|$ be the annotated and predicted link counts, $|\Pi|$ the number of matched pairs, $W$ the total weight $w$ of the matched pairs, and $\textit{Credit}$ the weight of matched pairs with agreeing labels, which is $\textit{Credit}_S$ for joint F1.
The unmatched-reference rate is $(|Y|-|\Pi|)/|Y|$, and the matched-prediction rate is $|\Pi|/|\widehat{Y}|$.
Conditional label accuracy counts correct labels among matched pairs, with faithful and unfaithful references also reported separately.
Correcting every matched label while retaining endpoints and the matching gives $2W/(|\widehat{Y}|+|Y|)$, the label-oracle F1.

The credit breakdown partitions $|Y|$ into correctly labeled credit $\textit{Credit}$, partial-overlap loss $|\Pi|-W$, label loss $W-\textit{Credit}$, and unmatched references $|Y|-|\Pi|$.
Partial-overlap loss measures incomplete overlap in matched pairs, and label loss is the weight of matched pairs whose labels disagree.
Thus, the unmatched share of missing recall credit is $(|Y|-|\Pi|)/(|Y|-\textit{Credit})$.

\noindent\textbf{Causes of unmatched links.}
For each unmatched annotated link, we inspect all predictions from the same response.
An endpoint is covered when a prediction overlaps its selected characters in the same document; an empty endpoint requires an empty prediction endpoint.
We first identify \emph{overmerged links}: an unmatched annotated link overlaps a predicted link at both endpoints, and that prediction has been matched to another annotated link.
For the remaining links, \emph{both sides missed} means neither endpoint is covered; \emph{counterpart missed} means only one is covered; and \emph{disconnected endpoints} means both are covered by separate predictions, with no prediction connecting them.
These four categories partition the unmatched links. Table~\ref{tab:unmatched-causes} further splits counterpart missed by the side of the missing endpoint.
The overlap criterion is identical to primary-match eligibility, so coverage can consist of a fragment of an endpoint.

\subsection{Prediction Splitting and Reference Grouping}
Both alternatives use the same content trimming and trials as the primary scores.
Eligible pairs must agree on empty endpoints and overlap at every nonempty endpoint.
Assignments maximize Dice/geometric overlap independently of labels.

\noindent\textbf{Prediction splitting.}
Each prediction selects the eligible reference with maximum Dice/geometric overlap; ties follow reference order.
Precision credit is the geometric mean of directed character precision at its endpoints.
For each reference, recall credit is computed against the endpoint unions of its assigned predictions.
For labeled prediction-splitting scores, each prediction earns credit only when its label agrees with its reference, and recall unions include only same-label predictions.

\noindent\textbf{Reference grouping.}
Reference grouping reverses this assignment: each reference selects its best eligible prediction, with ties following prediction order.
Recall credit is computed for each reference against its selected prediction; precision credit for each prediction uses the endpoint unions of its assigned references.
For full-context reference grouping, a group is unfaithful if any assigned reference is unfaithful, and faithful otherwise.
A prediction whose label disagrees with its group receives zero precision credit, and every reference in that group receives zero recall credit.
When the group label agrees, joint precision uses the union of all assigned reference endpoints and joint recall credits every member's geometric coverage, including faithful members of an unfaithful group.
The original prediction and reference counts remain the precision and recall denominators.
For defect F1, precision uses unfaithful predictions and their correctly labeled groups' full endpoint unions; recall uses only the original unfaithful reference links covered by correctly labeled unfaithful groups.
Thus faithful members contribute to an unfaithful group's precision coverage, while the defect-recall denominator continues to count annotated defects.
Unassigned links receive zero credit, and endpoint unions preserve document identity and gaps between spans.
Precision and recall are aggregated separately before taking their harmonic mean.

\section{Additional Results}
\label{app:additional-results}

\subsection{Defect Identification Across Repeated Attempts}
Most defective items remain incompletely diagnosed across three attempts (Table~\ref{tab:repetition-results}).
The highest complete-defect pass@3 is 40.2\%; even selecting the best response for each item leaves the remaining items with missed defects.
The repeated omissions are substantial: on average, 64.0\% of defects are never identified in any of a model's three attempts.
Complete identification of every link is rarer still, with an all-links pass@3 of at most 6.1\%.

\begin{table}[htbp]
    \centering
    \setlength{\tabcolsep}{3pt}
    \renewcommand{\arraystretch}{1.12}
    \caption{Full-context alignment F1 and complete-item pass@3 (\%) across three attempts, with single-attempt F1 reported as mean $\pm$ standard deviation.}
    \label{tab:repetition-results}
    \small
    \begin{tabular*}{\textwidth}{@{}@{\extracolsep{\fill}}lrrrrrr@{}}
        \toprule
        & \multicolumn{2}{c}{Defect F1} & \multicolumn{2}{c}{Joint F1} & \multicolumn{2}{c}{Complete-item pass@3} \\
        \cmidrule(lr){2-3}\cmidrule(lr){4-5}\cmidrule(lr){6-7}
        Model & Single & Best of 3 & Single & Best of 3 & All defects & All links \\
        \midrule
        GPT-5.6 Sol & $42.7\%\,{\scriptstyle\pm\,1.1\%}$ & 52.4\% & $45.5\%\,{\scriptstyle\pm\,0.1\%}$ & 51.0\% & 39.5\% & 5.7\% \\
Opus 5 & $31.9\%\,{\scriptstyle\pm\,0.2\%}$ & 39.3\% & $44.6\%\,{\scriptstyle\pm\,0.3\%}$ & 49.5\% & 27.7\% & 6.1\% \\
GPT-6 Astra & $47.8\%\,{\scriptstyle\pm\,0.5\%}$ & 52.6\% & $47.6\%\,{\scriptstyle\pm\,0.4\%}$ & 50.9\% & 37.8\% & 5.4\% \\
Opus 4.8 & $26.4\%\,{\scriptstyle\pm\,0.5\%}$ & 33.0\% & $44.3\%\,{\scriptstyle\pm\,0.1\%}$ & 49.0\% & 23.3\% & 5.7\% \\
Kimi K3 & $38.7\%\,{\scriptstyle\pm\,2.3\%}$ & 51.4\% & $44.1\%\,{\scriptstyle\pm\,0.8\%}$ & 50.9\% & 40.2\% & 5.9\% \\
GLM-5 & $13.5\%\,{\scriptstyle\pm\,0.9\%}$ & 23.6\% & $28.6\%\,{\scriptstyle\pm\,0.4\%}$ & 37.6\% & 18.6\% & 3.5\% \\
DeepSeek V3.2 & $18.7\%\,{\scriptstyle\pm\,0.7\%}$ & 28.1\% & $28.4\%\,{\scriptstyle\pm\,0.6\%}$ & 38.8\% & 20.6\% & 5.2\% \\
MiniMax M2.5 & $7.1\%\,{\scriptstyle\pm\,0.7\%}$ & 13.9\% & $12.4\%\,{\scriptstyle\pm\,0.2\%}$ & 19.5\% & 12.2\% & 0.0\% \\

        \bottomrule
    \end{tabular*}
\end{table}

\FloatBarrier
\subsection{Defects Involving Context}
As Figure~\ref{fig:context-defect-role} shows, 74.9\% of defects in context links lack a matched prediction across all models, and the share is 68.6\% for GPT-6 Astra.
Defects in links between formal context and a target or empty endpoint are especially likely to go unaligned: 94.1\% lack a matched prediction (Figure~\ref{fig:context-defect-endpoint}).
The informal-context group likewise connects source context to a target or empty endpoint.
Missing alignments also dominate failures in the other context groups.
These omissions leave defects in the supplied dependencies unlocated.

\begin{figure}[t]
    \centering
    \input{figures/context-defects}
    \caption{Defect outcomes by link role and context endpoint.}
    \label{fig:context-defects}
    \vspace{1.5em}
    \nextfloat
    \includegraphics[width=\linewidth]{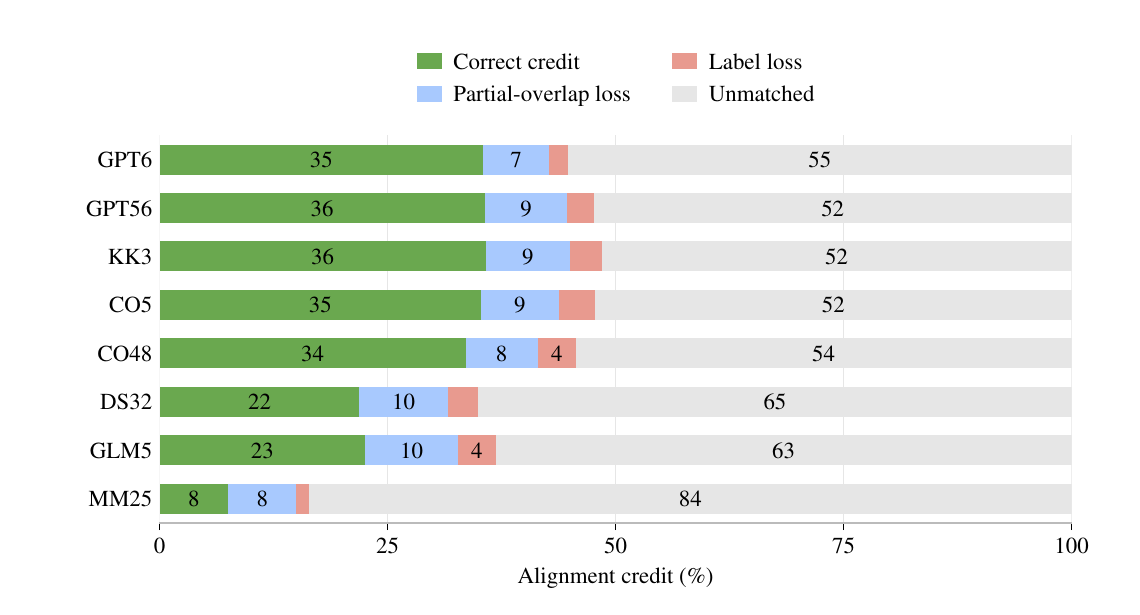}
    \caption{Alignment credit and losses by model, as percentages of annotated links.}
    \label{fig:recovery-diagnostics}
\end{figure}

\FloatBarrier
\subsection{Recognizing Underspecification}
\label{app:contextgap-results}
For all eight models, Table~\ref{tab:contextgap-paired-results} reports context-omitted alignment F1 on all 146 \contextgap{} items, together with verdicts on its 85 undecidable items and on their original items under full-context alignment.

\begin{table}[t]
    \centering
    \setlength{\tabcolsep}{3pt}
    \renewcommand{\arraystretch}{1.12}
    \caption{\contextgap{} scores and paired context judgments (\%, mean $\pm$ standard deviation across three repetitions).}
    \label{tab:contextgap-paired-results}
    \small
    \begin{tabular*}{\textwidth}{@{}@{\extracolsep{\fill}}lrrrrr@{}}
        \toprule
        & \multicolumn{3}{c}{\contextgap{} link F1} & \multicolumn{2}{c}{Paired item judgments} \\
        \cmidrule(lr){2-4}\cmidrule(lr){5-6}
        Model & F & U & ND & \shortstack{Gap: ND\\recognized} & \shortstack{Full context:\\correct} \\
        \midrule
        GPT-6 Astra & $49.2\%\,{\scriptstyle\pm\,0.1\%}$ & $34.2\%\,{\scriptstyle\pm\,0.8\%}$ & $17.4\%\,{\scriptstyle\pm\,0.1\%}$ & $28.2\%\,{\scriptstyle\pm\,2.0\%}$ & $96.9\%\,{\scriptstyle\pm\,0.7\%}$ \\
GPT-5.6 Sol & $47.3\%\,{\scriptstyle\pm\,0.8\%}$ & $32.0\%\,{\scriptstyle\pm\,2.1\%}$ & $11.5\%\,{\scriptstyle\pm\,0.6\%}$ & $16.5\%\,{\scriptstyle\pm\,1.2\%}$ & $93.7\%\,{\scriptstyle\pm\,1.8\%}$ \\
Kimi K3 & $45.3\%\,{\scriptstyle\pm\,2.1\%}$ & $31.8\%\,{\scriptstyle\pm\,1.9\%}$ & $5.9\%\,{\scriptstyle\pm\,0.6\%}$ & $8.6\%\,{\scriptstyle\pm\,0.7\%}$ & $94.9\%\,{\scriptstyle\pm\,1.4\%}$ \\
Opus 5 & $44.2\%\,{\scriptstyle\pm\,0.2\%}$ & $29.2\%\,{\scriptstyle\pm\,0.8\%}$ & $1.7\%\,{\scriptstyle\pm\,0.8\%}$ & $2.4\%\,{\scriptstyle\pm\,2.0\%}$ & $94.9\%\,{\scriptstyle\pm\,0.7\%}$ \\
Opus 4.8 & $44.2\%\,{\scriptstyle\pm\,0.5\%}$ & $26.4\%\,{\scriptstyle\pm\,2.7\%}$ & $0.4\%\,{\scriptstyle\pm\,0.5\%}$ & $1.6\%\,{\scriptstyle\pm\,0.7\%}$ & $91.4\%\,{\scriptstyle\pm\,0.7\%}$ \\
DeepSeek V3.2 & $23.9\%\,{\scriptstyle\pm\,1.8\%}$ & $12.9\%\,{\scriptstyle\pm\,1.6\%}$ & $4.3\%\,{\scriptstyle\pm\,0.9\%}$ & $25.9\%\,{\scriptstyle\pm\,5.1\%}$ & $63.1\%\,{\scriptstyle\pm\,5.8\%}$ \\
GLM-5 & $33.4\%\,{\scriptstyle\pm\,0.4\%}$ & $9.9\%\,{\scriptstyle\pm\,3.8\%}$ & $2.9\%\,{\scriptstyle\pm\,0.6\%}$ & $14.5\%\,{\scriptstyle\pm\,0.7\%}$ & $65.1\%\,{\scriptstyle\pm\,2.4\%}$ \\
MiniMax M2.5 & $21.7\%\,{\scriptstyle\pm\,1.1\%}$ & $7.4\%\,{\scriptstyle\pm\,1.8\%}$ & $0.3\%\,{\scriptstyle\pm\,0.2\%}$ & $1.6\%\,{\scriptstyle\pm\,0.7\%}$ & $82.4\%\,{\scriptstyle\pm\,4.1\%}$ \\

        \bottomrule
    \end{tabular*}
\end{table}

\noindent\textbf{Persistence of definite judgments.}
Across three responses per undecidable item, Astra gives 65 faithful, 118 unfaithful, and 72 undecidable judgments.
It never recognizes undecidability on 55 of the 85 items; 52 of these also receive correct judgments for both full-context interpretations in all three repetitions.
Twenty items are recognized as undecidable in all three responses, eight in one response, and two in two responses.
These repeated judgments show how frequently the model settles on a definite answer despite the missing information.

\begingroup
\lstdefinelanguage{FaithAlignLean}{
    morekeywords={abbrev,axiom,by,def,end,fun,import,in,inductive,instance,
      namespace,noncomputable,notation,opaque,open,section,structure,theorem,
      universe,variable,where,match,with,if,then,else,let,have,show,exact,
      refine,cases,induction,sorry,syntax,macro_rules,macro,infixl},
    sensitive=true,
    morecomment=[l]{--},
    morecomment=[n]{/-}{-/},
    morestring=[b]"
}
\lstdefinestyle{contextgapLean}{
    language=FaithAlignLean,
    basicstyle=\small\ttfamily,
    keywordstyle=\bfseries\color{blue!45!black},
    commentstyle=\color{black!60},
    stringstyle=\color{black},
    columns=fullflexible,
    keepspaces=true,
    showstringspaces=false,
    breaklines=true,
    breakatwhitespace=true,
    breakindent=1em,
    tabsize=2,
    frame=single,
    rulecolor=\color{black!25},
    framerule=.3pt,
    framesep=4pt,
    xleftmargin=5pt,
    xrightmargin=5pt,
    aboveskip=6pt,
    belowskip=6pt,
    upquote=true,
    literate=
        {¬}{{\ensuremath{\neg}}}1
        {·}{{\ensuremath{\cdot}}}1
        {×}{{\ensuremath{\times}}}1
        {Γ}{{\ensuremath{\Gamma}}}1
        {Δ}{{\ensuremath{\Delta}}}1
        {Θ}{{\ensuremath{\Theta}}}1
        {Λ}{{\ensuremath{\Lambda}}}1
        {α}{{\ensuremath{\alpha}}}1
        {β}{{\ensuremath{\beta}}}1
        {ι}{{\ensuremath{\iota}}}1
        {λ}{{\ensuremath{\lambda}}}1
        {₁}{{\ensuremath{{}_{1}}}}1
        {₂}{{\ensuremath{{}_{2}}}}1
        {ℕ}{{\ensuremath{\mathbb{N}}}}1
        {ℝ}{{\ensuremath{\mathbb{R}}}}1
        {→}{{\ensuremath{\to}}}1
        {∀}{{\ensuremath{\forall}}}1
        {∃}{{\ensuremath{\exists}}}1
        {∅}{{\ensuremath{\varnothing}}}1
        {∈}{{\ensuremath{\in}}}1
        {∉}{{\ensuremath{\notin}}}1
        {∟}{{\ensuremath{\llcorner}}}1
        {∠}{{\ensuremath{\angle}}}1
        {∧}{{\ensuremath{\land}}}1
        {∨}{{\ensuremath{\lor}}}1
        {∪}{{\ensuremath{\cup}}}1
        {≠}{{\ensuremath{\ne}}}1
        {≤}{{\ensuremath{\le}}}1
        {≥}{{\ensuremath{\ge}}}1
        {─}{{\ensuremath{\rule[.5ex]{.55em}{.04em}}}}1
        {△}{{\ensuremath{\triangle}}}1
        {⟨}{{\ensuremath{\langle}}}1
        {⟩}{{\ensuremath{\rangle}}}1
        {⬝}{{\ensuremath{\vcenter{\hbox{\rule{.2ex}{.2ex}}}}}}1
}

\subsubsection{\contextgap{} Examples}

\label{app:contextgap-examples}

The input excerpts below use $T_N$, $T_F$, $C_N$, $I$, and $C_F$. Each triplet shares its targets and formal context and pairs the original item with a constructed counterpart of the opposite verdict. The tables specify the informal context and assumptions, annotated labels, and recorded Astra judgments. F, U, and ND denote faithful, unfaithful, and undecidable; $\varnothing$ denotes an absent component.

\noindent\textbf{Example 1: Isosceles triangles (LeanEuclid--Pistis 1.05; \citealp{fitzpatrick2008euclid}).}
Both versions state an authored assumption.
The faithful version adopts the inclusive convention, under which equilateral triangles are isosceles, while Euclid's definition excludes them.
Both conventions are in use, and each version is judged under the assumption it states.

\begin{center}

\small

\setlength{\tabcolsep}{5pt}

\begin{tabular*}{\linewidth}{@{}@{\extracolsep{\fill}}lcccc@{}}

\toprule

Input & $C_N$ & $I$ & Label & Astra \\

\midrule

Full F & $\varnothing$ & $I^{\mathrm{F}}$ & F & F, F, F \\

Full U & $\varnothing$ & $I^{\mathrm{U}}$ & U & U, U, U \\

\contextgap{} & $\varnothing$ & $\varnothing$ & ND & F, F, F \\

\bottomrule

\end{tabular*}

\end{center}

\noindent\textbf{$T_N$: Natural Language Target Statement.}

For isosceles triangles, the angles at the base are equal to one another, and if the equal sides are produced then the angles under the base will be equal to one another. Let $ABC$ be an isosceles triangle having the side $AB$ equal to the side $AC$, and let the straight-lines $BD$ and $CE$ have been produced in a straight-line with $AB$ and $AC$ (respectively) [Post.~2]. I say that the angle $ABC$ is equal to $ACB$, and (angle) $CBD$  to $BCE$.

\par\medskip\noindent\textbf{$T_F$: Formal Lean Target Statement.}\par

\begin{lstlisting}[style=contextgapLean]
theorem proposition_5 : ∀ (a b c d e : Point) (AB BC AC : Line),
  formTriangle a b c AB BC AC ∧ (|(a─b)| = |(a─c)|) ∧
  (between a b d) ∧ (between a c e) →
  (∠ a:b:c = ∠ a:c:b) ∧ (∠ c:b:d = ∠ b:c:e) := by sorry
\end{lstlisting}

\noindent\textbf{$I^{\mathrm F}$: Implicit Assumption in Natural Language (Faithful).}

Isosceles means at least two equal sides, including equilateral triangles.

\noindent\textbf{$I^{\mathrm U}$: Implicit Assumption in Natural Language (Unfaithful).}

Isosceles means exactly two equal sides; equilateral triangles are excluded.

\noindent\textbf{Example 2: Strictly upper triangular matrices (MA-072; \citealp{lang2002algebra}).}

\begin{center}

\small

\setlength{\tabcolsep}{5pt}

\begin{tabular*}{\linewidth}{@{}@{\extracolsep{\fill}}lcccc@{}}

\toprule

Input & $C_N$ & $I$ & Label & Astra \\

\midrule

Full F & $\varnothing$ & $I^{\mathrm{F}}$ & F & F, F, F \\

Full U & $C_N^{\mathrm{U}}$ & $\varnothing$ & U & U, U, U \\

\contextgap{} & $\varnothing$ & $\varnothing$ & ND & ND, ND, ND \\

\bottomrule

\end{tabular*}

\end{center}

\noindent\textbf{$T_N$: Natural Language Target Statement.}

Let $N$ be a strictly upper triangular $n\times n$ matrix, that is $N=(a_{ij})$ and $a_{ij}=0$ if $i\geqq j$. Show that $N^{n}=0$.

\par\medskip\noindent\textbf{$T_F$: Formal Lean Target Statement.}\par

\begin{lstlisting}[style=contextgapLean]
theorem nilpotent_matrix_power (n : ℕ) (N : Matrix (Fin n) (Fin n) ℝ)
  (h : ∀ i j : Fin n, i ≥ j → N i j = 0) :
  N ^ n = 0 := by sorry
\end{lstlisting}

\noindent\textbf{$C_N^{\mathrm U}$: Natural Language Context (Unfaithful).}

Source convention (Lang, Algebra, Chapter XIII, p. 503): matrix coefficients range over a commutative ring R. Exercise 8 on p. 546 is stated under this convention.

\noindent\textbf{$I^{\mathrm F}$: Implicit Assumption in Natural Language (Faithful).}

All matrices in this exercise have real entries.

\noindent\textbf{Example 3: The cut rule (LCS-Bench lcs-003; \citealp{gallier2015logic}).}

\begin{center}

\small

\setlength{\tabcolsep}{5pt}

\begin{tabular*}{\linewidth}{@{}@{\extracolsep{\fill}}lcccc@{}}

\toprule

Input & $C_N$ & $I$ & Label & Astra \\

\midrule

Full F & $C_N^{\mathrm{F}}$ & $\varnothing$ & F & F, F, F \\

Full U & $C_N^{\mathrm{U}}$ & $I^{\mathrm{U}}$ & U & U, U, U \\

\contextgap{} & $\varnothing$ & $\varnothing$ & ND & U, U, U \\

\bottomrule

\end{tabular*}

\end{center}

\noindent\textbf{$T_N$: Natural Language Target Statement.}

The sequents of the system \(G1^{nnf}\) are pairs \(\Gamma \to \Delta\), where \(\Gamma\) and \(\Delta\) are finite \emph{sets} of formulae in NNF.

The cut rule of \(G1^{nnf}\) infers \(\Gamma, \Lambda \to \Delta, \Theta\) from the premises \(\Gamma \to \Delta, A\) and \(A, \Lambda \to \Theta\). The formula \(A\) is called the cut formula of this inference.

\par\medskip\noindent\textbf{$T_F$: Formal Lean Target Statement.}\par

\begin{lstlisting}[style=contextgapLean]
def ApplicationConditions {L : Language}
    (rule : G1nnfCutRule L) (Γ : LCS.Finset (FormulaN L))
    (Δ : LCS.Finset (FormulaN L)) (Λ : LCS.Finset (FormulaN L))
    (Θ : LCS.Finset (FormulaN L)) : Prop :=
  G1nnfFinsetSequent.IsNNF (rule.conclusion Γ Δ Λ Θ) ∧
  G1nnfFinsetSequent.NoFreeBound (rule.conclusion Γ Δ Λ Θ) ∧
  ∀ s ∈ rule.premises Γ Δ Λ Θ,
    G1nnfFinsetSequent.IsNNF s ∧ G1nnfFinsetSequent.NoFreeBound s
\end{lstlisting}

\noindent\textbf{$C_N^{\mathrm F}$: Natural Language Context (Faithful).}

Predicate symbols: A (countable, possibly empty) set PS of symbols \(P_{0}, P_{1}, \ldots\), and a rank function \(r\) assigning a nonnegative integer \(r(P)\) (called rank or arity) to each predicate symbol \(P\).

\noindent\textbf{$C_N^{\mathrm F}$ and $C_N^{\mathrm U}$: Natural Language Context (Faithful and Unfaithful).}

First, we consider the case of a first-order language without equality. In this Chapter, it is assumed that no variable occurs both free and bound in any sequent (or formula).

\noindent\textbf{$I^{\mathrm U}$: Implicit Assumption in Natural Language (Unfaithful).}

PS may be uncountable.

\par\medskip\noindent\textbf{$C_F$: Formal Lean Context.}\par

\begin{lstlisting}[style=contextgapLean]
structure Language : Type 1 where
  -- ...
  predSymbols : Type
  predArity : predSymbols → Nat
  hasEquality : Bool := false
  equalityFree : hasEquality = false
  -- ...
  predSymbolsCountable : ∃ code : predSymbols → Nat, ∀ a b, code a = code b → a = b
  -- ...
\end{lstlisting}

\begin{lstlisting}[style=contextgapLean]
def NoFreeBound {L : Language}
    (s : LCS.Finset (FormulaN L) × LCS.Finset (FormulaN L)) : Prop :=
  ∀ A, A ∈ s.1 ∨ A ∈ s.2 →
    ∀ B, B ∈ s.1 ∨ B ∈ s.2 →
      ∀ x, x ∈ Variables.formulaFVN A → x ∈ Variables.formulaBVN B → False
\end{lstlisting}

\endgroup

\FloatBarrier
\subsection{Fine-grained Alignment and Labeling}
Missing links account for most lost alignment credit (Figure~\ref{fig:recovery-diagnostics}).
Correcting every label at the existing predicted endpoints raises joint F1 by 4.2 percentage points on average (Table~\ref{tab:recovery-results}).
Most of the missing credit consequently remains tied to endpoints and absent links.

\begin{table}[htbp]
    \centering
    \setlength{\tabcolsep}{3pt}
    \renewcommand{\arraystretch}{1.12}
    \caption{Alignment coverage, conditional labeling accuracy, and observed and label-oracle joint F1 (\%), as mean $\pm$ standard deviation across three repetitions.}
    \label{tab:recovery-results}
    \small
    \begin{tabular*}{\textwidth}{@{}@{\extracolsep{\fill}}lrrrrrr@{}}
        \toprule
        & Annotated & Predicted & \multicolumn{2}{c}{Conditional labeling} & \multicolumn{2}{c}{Joint F1} \\
        \cmidrule(lr){4-5}\cmidrule(lr){6-7}
        Model & unmatched & matched & Faithful & Unfaithful & Observed & Oracle \\
        \midrule
        GPT-5.6 Sol & $52.4\%\,{\scriptstyle\pm\,0.1\%}$ & $83.8\%\,{\scriptstyle\pm\,0.5\%}$ & $97.4\%\,{\scriptstyle\pm\,0.3\%}$ & $71.3\%\,{\scriptstyle\pm\,0.6\%}$ & $45.5\%\,{\scriptstyle\pm\,0.1\%}$ & $49.2\%\,{\scriptstyle\pm\,0.1\%}$ \\
Opus 5 & $52.3\%\,{\scriptstyle\pm\,0.2\%}$ & $82.3\%\,{\scriptstyle\pm\,0.7\%}$ & $98.6\%\,{\scriptstyle\pm\,0.1\%}$ & $50.1\%\,{\scriptstyle\pm\,1.3\%}$ & $44.6\%\,{\scriptstyle\pm\,0.3\%}$ & $49.6\%\,{\scriptstyle\pm\,0.3\%}$ \\
GPT-6 Astra & $55.2\%\,{\scriptstyle\pm\,0.3\%}$ & $91.4\%\,{\scriptstyle\pm\,0.2\%}$ & $96.7\%\,{\scriptstyle\pm\,0.2\%}$ & $82.2\%\,{\scriptstyle\pm\,0.1\%}$ & $47.6\%\,{\scriptstyle\pm\,0.4\%}$ & $50.5\%\,{\scriptstyle\pm\,0.3\%}$ \\
Opus 4.8 & $54.4\%\,{\scriptstyle\pm\,0.1\%}$ & $88.9\%\,{\scriptstyle\pm\,0.3\%}$ & $99.2\%\,{\scriptstyle\pm\,0.1\%}$ & $39.8\%\,{\scriptstyle\pm\,0.4\%}$ & $44.3\%\,{\scriptstyle\pm\,0.1\%}$ & $49.9\%\,{\scriptstyle\pm\,0.1\%}$ \\
Kimi K3 & $51.5\%\,{\scriptstyle\pm\,0.8\%}$ & $77.9\%\,{\scriptstyle\pm\,0.6\%}$ & $97.8\%\,{\scriptstyle\pm\,0.4\%}$ & $61.8\%\,{\scriptstyle\pm\,3.8\%}$ & $44.1\%\,{\scriptstyle\pm\,0.8\%}$ & $48.3\%\,{\scriptstyle\pm\,0.4\%}$ \\
GLM-5 & $63.1\%\,{\scriptstyle\pm\,0.5\%}$ & $63.9\%\,{\scriptstyle\pm\,0.2\%}$ & $93.1\%\,{\scriptstyle\pm\,0.4\%}$ & $36.3\%\,{\scriptstyle\pm\,1.6\%}$ & $28.6\%\,{\scriptstyle\pm\,0.4\%}$ & $33.8\%\,{\scriptstyle\pm\,0.4\%}$ \\
DeepSeek V3.2 & $65.0\%\,{\scriptstyle\pm\,0.3\%}$ & $64.6\%\,{\scriptstyle\pm\,2.7\%}$ & $92.8\%\,{\scriptstyle\pm\,0.3\%}$ & $52.8\%\,{\scriptstyle\pm\,2.0\%}$ & $28.4\%\,{\scriptstyle\pm\,0.6\%}$ & $32.8\%\,{\scriptstyle\pm\,0.6\%}$ \\
MiniMax M2.5 & $83.6\%\,{\scriptstyle\pm\,0.0\%}$ & $77.9\%\,{\scriptstyle\pm\,3.0\%}$ & $96.2\%\,{\scriptstyle\pm\,0.2\%}$ & $32.2\%\,{\scriptstyle\pm\,4.4\%}$ & $12.4\%\,{\scriptstyle\pm\,0.2\%}$ & $14.7\%\,{\scriptstyle\pm\,0.3\%}$ \\

        \bottomrule
    \end{tabular*}
\end{table}

\FloatBarrier
\noindent\textbf{Why links remain unmatched.}
An unmatched link can arise even when the response mentions both endpoints.
The model may place them in separate predictions, or combine several annotated links into one broad prediction matched to another link.
Table~\ref{tab:unmatched-causes} separates these cases from missing one or both endpoints.
These categories distinguish missing mathematical content from missing connections between selected regions.

\begin{table}[htbp]
    \centering
    \setlength{\tabcolsep}{3pt}
    \renewcommand{\arraystretch}{1.12}
    \caption{Unmatched links ($n$) and failure-type percentages across three repetitions, using the categories in Figure~\ref{fig:unmatched-links}.}
    \label{tab:unmatched-causes}
    \small
    \begin{tabular*}{\textwidth}{@{}@{\extracolsep{\fill}}lrrrrrr@{}}
        \toprule
        Model & $n$ & \shortstack{Both\\missed} & \shortstack{Formal\\missed} & \shortstack{Informal\\missed} & Disconnected & Overmerged \\
        \midrule
        GPT-5.6 Sol & 7493 & 14.4\% & 9.6\% & 16.3\% & 14.7\% & 45.0\% \\
Opus 5 & 7474 & 20.5\% & 10.9\% & 17.7\% & 21.2\% & 29.7\% \\
GPT-6 Astra & 7894 & 16.5\% & 10.7\% & 20.8\% & 11.1\% & 40.8\% \\
Opus 4.8 & 7776 & 25.1\% & 12.6\% & 19.3\% & 12.9\% & 30.1\% \\
Kimi K3 & 7370 & 17.9\% & 8.5\% & 15.4\% & 13.7\% & 44.4\% \\
GLM-5 & 9032 & 20.7\% & 13.0\% & 22.8\% & 19.1\% & 24.5\% \\
DeepSeek V3.2 & 9304 & 39.0\% & 10.5\% & 15.3\% & 16.1\% & 19.1\% \\
MiniMax M2.5 & 11958 & 34.2\% & 15.7\% & 20.6\% & 1.9\% & 27.6\% \\

        \bottomrule
    \end{tabular*}
\end{table}

\FloatBarrier
\noindent\textbf{Allowing finer or broader predictions.}
Prediction splitting permits several predictions to describe one annotated link; reference grouping permits one prediction to describe several annotated links.
Table~\ref{tab:grouping-results} reports both alternatives, using the scoring definitions in Appendix~\ref{app:metric-details}.
Defect F1 under reference grouping is 33.1\% on average.
Defect identification remains weak when one prediction can cover several annotated links.
Complete identification remains low under reference grouping, where a defect counts as identified when the prediction it selects is labeled unfaithful.
GPT-6 Astra then identifies every defect in 39.2\% of defective items and in 9.8\% of items with two or more defects, against 34.3\% and 4.5\% under standard scoring.

\begin{table}[htbp]
    \centering
    \setlength{\tabcolsep}{3pt}
    \renewcommand{\arraystretch}{1.12}
    \caption{Full-context alignment F1 (\%) under standard scoring, prediction splitting, and reference grouping, as mean $\pm$ standard deviation across three repetitions.}
    \label{tab:grouping-results}
    \small
    \begin{tabular*}{\textwidth}{@{}@{\extracolsep{\fill}}lrrrrrr@{}}
        \toprule
        & \multicolumn{2}{c}{Standard} & \multicolumn{2}{c}{Prediction splitting} & \multicolumn{2}{c}{Reference grouping} \\
        \cmidrule(lr){2-3}\cmidrule(lr){4-5}\cmidrule(lr){6-7}
        Model & Defect & Joint & Defect & Joint & Defect & Joint \\
        \midrule
        GPT-5.6 Sol & $42.7\%\,{\scriptstyle\pm\,1.1\%}$ & $45.5\%\,{\scriptstyle\pm\,0.1\%}$ & $45.9\%\,{\scriptstyle\pm\,0.7\%}$ & $49.1\%\,{\scriptstyle\pm\,0.1\%}$ & $49.3\%\,{\scriptstyle\pm\,1.1\%}$ & $63.7\%\,{\scriptstyle\pm\,0.3\%}$ \\
Opus 5 & $31.9\%\,{\scriptstyle\pm\,0.2\%}$ & $44.6\%\,{\scriptstyle\pm\,0.3\%}$ & $33.1\%\,{\scriptstyle\pm\,0.2\%}$ & $48.2\%\,{\scriptstyle\pm\,0.3\%}$ & $35.7\%\,{\scriptstyle\pm\,0.3\%}$ & $56.5\%\,{\scriptstyle\pm\,0.3\%}$ \\
GPT-6 Astra & $47.8\%\,{\scriptstyle\pm\,0.5\%}$ & $47.6\%\,{\scriptstyle\pm\,0.4\%}$ & $49.6\%\,{\scriptstyle\pm\,0.6\%}$ & $50.4\%\,{\scriptstyle\pm\,0.4\%}$ & $55.8\%\,{\scriptstyle\pm\,0.7\%}$ & $67.5\%\,{\scriptstyle\pm\,0.3\%}$ \\
Opus 4.8 & $26.4\%\,{\scriptstyle\pm\,0.5\%}$ & $44.3\%\,{\scriptstyle\pm\,0.1\%}$ & $27.0\%\,{\scriptstyle\pm\,0.4\%}$ & $47.6\%\,{\scriptstyle\pm\,0.2\%}$ & $29.7\%\,{\scriptstyle\pm\,0.5\%}$ & $57.2\%\,{\scriptstyle\pm\,0.4\%}$ \\
Kimi K3 & $38.7\%\,{\scriptstyle\pm\,2.3\%}$ & $44.1\%\,{\scriptstyle\pm\,0.8\%}$ & $41.0\%\,{\scriptstyle\pm\,2.5\%}$ & $47.5\%\,{\scriptstyle\pm\,0.9\%}$ & $45.1\%\,{\scriptstyle\pm\,2.0\%}$ & $60.5\%\,{\scriptstyle\pm\,0.7\%}$ \\
GLM-5 & $13.5\%\,{\scriptstyle\pm\,0.9\%}$ & $28.6\%\,{\scriptstyle\pm\,0.4\%}$ & $14.8\%\,{\scriptstyle\pm\,0.9\%}$ & $32.6\%\,{\scriptstyle\pm\,0.4\%}$ & $17.0\%\,{\scriptstyle\pm\,0.4\%}$ & $39.2\%\,{\scriptstyle\pm\,0.2\%}$ \\
DeepSeek V3.2 & $18.7\%\,{\scriptstyle\pm\,0.7\%}$ & $28.4\%\,{\scriptstyle\pm\,0.6\%}$ & $19.8\%\,{\scriptstyle\pm\,0.6\%}$ & $31.9\%\,{\scriptstyle\pm\,0.6\%}$ & $22.6\%\,{\scriptstyle\pm\,0.9\%}$ & $38.3\%\,{\scriptstyle\pm\,0.7\%}$ \\
MiniMax M2.5 & $7.1\%\,{\scriptstyle\pm\,0.7\%}$ & $12.4\%\,{\scriptstyle\pm\,0.2\%}$ & $7.3\%\,{\scriptstyle\pm\,0.8\%}$ & $15.6\%\,{\scriptstyle\pm\,0.1\%}$ & $10.0\%\,{\scriptstyle\pm\,1.5\%}$ & $32.1\%\,{\scriptstyle\pm\,0.6\%}$ \\

        \bottomrule
    \end{tabular*}
\end{table}

Grouping also leaves many annotated links unaligned (Table~\ref{tab:grouped-outcomes}): 40.7\% remain unmatched, and 7.8\% belong to incorrectly labeled groups.
A correctly labeled group can still receive partial overlap credit when its endpoints cover only part of the assigned content.

\begin{table}[htbp]
    \centering
    \setlength{\tabcolsep}{3pt}
    \renewcommand{\arraystretch}{1.12}
    \caption{Link outcome counts (percentages) under reference grouping across three repetitions.}
    \label{tab:grouped-outcomes}
    \small
    \begin{tabular*}{\textwidth}{@{}@{\extracolsep{\fill}}lrrrr@{}}
        \toprule
        Model & Total & Unmatched & Matched, wrong & Matched, correct \\
        \midrule
        GPT-5.6 Sol & 14,304 & 4,123 (28.8\%) & 910 (6.4\%) & 9,271 (64.8\%) \\
Opus 5 & 14,304 & 5,257 (36.8\%) & 1,083 (7.6\%) & 7,964 (55.7\%) \\
GPT-6 Astra & 14,304 & 4,670 (32.6\%) & 642 (4.5\%) & 8,992 (62.9\%) \\
Opus 4.8 & 14,304 & 5,437 (38.0\%) & 1,180 (8.2\%) & 7,687 (53.7\%) \\
Kimi K3 & 14,304 & 4,097 (28.6\%) & 1,055 (7.4\%) & 9,152 (64.0\%) \\
GLM-5 & 14,304 & 6,818 (47.7\%) & 1,480 (10.3\%) & 6,006 (42.0\%) \\
DeepSeek V3.2 & 14,304 & 7,531 (52.6\%) & 1,050 (7.3\%) & 5,723 (40.0\%) \\
MiniMax M2.5 & 14,304 & 8,654 (60.5\%) & 1,519 (10.6\%) & 4,131 (28.9\%) \\

        \bottomrule
    \end{tabular*}
\end{table}

\FloatBarrier
\subsection{Selected-region Counterpart Identification}
\label{sec:results-anchors}
This task supplies a region selected from one target and asks the model to identify all its counterparts and label each resulting link.
We evaluate both directions: informal-target regions mapped to formal regions, and formal-target regions mapped to informal regions.
The complete context remains available, and counterparts may lie in either targets or context.
A selected region can have multiple counterparts; an empty counterpart records unmatched material.
Supplying one endpoint isolates counterpart identification and labeling from the initial selection of a region.

Fixing the region of interest still leaves substantial difficulty identifying its complete counterpart set.
Only 24.7\% of selected regions have the exact complete counterpart set, including its boundaries.
The outputs omit counterparts or select their boundaries imprecisely, and defect F1 averages 32.4\% (Table~\ref{tab:selected-region-results}).

\begin{table}[htbp]
    \centering
    \setlength{\tabcolsep}{3pt}
    \renewcommand{\arraystretch}{1.12}
    \caption{Selected-region counterpart identification F1 (\%) in the two directions, with one repetition per model.}
    \label{tab:selected-region-results}
    \small
    \begin{tabular*}{\textwidth}{@{}@{\extracolsep{\fill}}lrrrr@{}}
        \toprule
        & \multicolumn{2}{c}{Informal region} & \multicolumn{2}{c}{Formal region} \\
        \cmidrule(lr){2-3}\cmidrule(lr){4-5}
        Model & Defect & Joint & Defect & Joint \\
        \midrule
        GPT-5.6 Sol & 45.6\% & 66.4\% & 45.6\% & 65.3\% \\
Opus 5 & 41.6\% & 67.8\% & 46.8\% & 67.1\% \\
GPT-6 Astra & 52.2\% & 69.3\% & 56.9\% & 69.0\% \\
Opus 4.8 & 33.8\% & 67.2\% & 32.6\% & 64.2\% \\
Kimi K3 & 33.2\% & 64.8\% & 38.5\% & 62.5\% \\
GLM-5 & 21.8\% & 49.4\% & 12.4\% & 45.0\% \\
DeepSeek V3.2 & 26.8\% & 51.2\% & 18.2\% & 44.0\% \\
MiniMax M2.5 & 6.2\% & 30.8\% & 6.1\% & 29.5\% \\

        \bottomrule
    \end{tabular*}
\end{table}

\FloatBarrier
\subsection{Results by Source}
Performance varies across the four source groups (Table~\ref{tab:source-results}).
Every model achieves its highest joint F1 on ProofNet-Verified.
Defect F1 follows a different pattern: for example, GPT-5.6 Sol scores highest on LeanEuclid--Pistis despite its lower joint F1 there.
The source composition in Appendix~\ref{app:dataset-details} helps explain why overall alignment and defect identification need separate evaluation: the sources differ in both link structure and the proportion of defective links.

\begin{table}[htbp]
    \centering
    \setlength{\tabcolsep}{3pt}
    \renewcommand{\arraystretch}{1.12}
    \caption{Full-context alignment F1 (\%) by source, as mean $\pm$ standard deviation across three repetitions.}
    \label{tab:source-results}
    \small
    \begin{tabular*}{\textwidth}{@{}@{\extracolsep{\fill}}lrrrr@{}}
        \toprule
        & \multicolumn{2}{c}{LCS-Bench ($n=196$)} & \multicolumn{2}{c}{ProofNet-Verified ($n=77$)} \\
        \cmidrule(lr){2-3}\cmidrule(lr){4-5}
        Model & Defect & Joint & Defect & Joint \\
        \midrule
        GPT-5.6 Sol & $44.0\%\,{\scriptstyle\pm\,1.9\%}$ & $50.9\%\,{\scriptstyle\pm\,0.5\%}$ & $32.9\%\,{\scriptstyle\pm\,4.3\%}$ & $58.2\%\,{\scriptstyle\pm\,1.0\%}$ \\
Opus 5 & $29.4\%\,{\scriptstyle\pm\,0.6\%}$ & $49.3\%\,{\scriptstyle\pm\,0.2\%}$ & $30.9\%\,{\scriptstyle\pm\,2.5\%}$ & $57.5\%\,{\scriptstyle\pm\,0.9\%}$ \\
GPT-6 Astra & $50.1\%\,{\scriptstyle\pm\,1.0\%}$ & $55.8\%\,{\scriptstyle\pm\,0.5\%}$ & $44.4\%\,{\scriptstyle\pm\,2.5\%}$ & $56.7\%\,{\scriptstyle\pm\,0.6\%}$ \\
Opus 4.8 & $22.2\%\,{\scriptstyle\pm\,0.9\%}$ & $49.1\%\,{\scriptstyle\pm\,0.5\%}$ & $29.5\%\,{\scriptstyle\pm\,1.6\%}$ & $61.4\%\,{\scriptstyle\pm\,1.5\%}$ \\
Kimi K3 & $38.8\%\,{\scriptstyle\pm\,5.5\%}$ & $48.4\%\,{\scriptstyle\pm\,1.6\%}$ & $36.8\%\,{\scriptstyle\pm\,1.2\%}$ & $57.8\%\,{\scriptstyle\pm\,0.9\%}$ \\
GLM-5 & $10.7\%\,{\scriptstyle\pm\,1.2\%}$ & $29.7\%\,{\scriptstyle\pm\,0.4\%}$ & $15.3\%\,{\scriptstyle\pm\,7.5\%}$ & $39.3\%\,{\scriptstyle\pm\,3.9\%}$ \\
DeepSeek V3.2 & $16.5\%\,{\scriptstyle\pm\,0.5\%}$ & $26.3\%\,{\scriptstyle\pm\,1.5\%}$ & $18.2\%\,{\scriptstyle\pm\,1.9\%}$ & $41.3\%\,{\scriptstyle\pm\,1.8\%}$ \\
MiniMax M2.5 & $6.9\%\,{\scriptstyle\pm\,1.0\%}$ & $11.4\%\,{\scriptstyle\pm\,0.3\%}$ & $10.3\%\,{\scriptstyle\pm\,5.5\%}$ & $25.5\%\,{\scriptstyle\pm\,0.5\%}$ \\

        \midrule
        & \multicolumn{2}{c}{MA-Align ($n=122$)} & \multicolumn{2}{c}{LeanEuclid--Pistis ($n=65$)} \\
        \cmidrule(lr){2-3}\cmidrule(lr){4-5}
        Model & Defect & Joint & Defect & Joint \\
        \midrule
        GPT-5.6 Sol & $42.5\%\,{\scriptstyle\pm\,0.9\%}$ & $35.9\%\,{\scriptstyle\pm\,0.3\%}$ & $53.2\%\,{\scriptstyle\pm\,1.1\%}$ & $34.2\%\,{\scriptstyle\pm\,1.1\%}$ \\
Opus 5 & $33.7\%\,{\scriptstyle\pm\,0.5\%}$ & $34.3\%\,{\scriptstyle\pm\,0.4\%}$ & $38.2\%\,{\scriptstyle\pm\,4.3\%}$ & $37.7\%\,{\scriptstyle\pm\,1.0\%}$ \\
GPT-6 Astra & $46.1\%\,{\scriptstyle\pm\,0.4\%}$ & $35.8\%\,{\scriptstyle\pm\,0.2\%}$ & $48.7\%\,{\scriptstyle\pm\,1.6\%}$ & $34.4\%\,{\scriptstyle\pm\,0.4\%}$ \\
Opus 4.8 & $27.7\%\,{\scriptstyle\pm\,1.5\%}$ & $31.2\%\,{\scriptstyle\pm\,0.5\%}$ & $49.0\%\,{\scriptstyle\pm\,4.3\%}$ & $39.6\%\,{\scriptstyle\pm\,0.1\%}$ \\
Kimi K3 & $38.3\%\,{\scriptstyle\pm\,2.0\%}$ & $34.4\%\,{\scriptstyle\pm\,1.1\%}$ & $47.0\%\,{\scriptstyle\pm\,3.3\%}$ & $35.0\%\,{\scriptstyle\pm\,1.7\%}$ \\
GLM-5 & $14.9\%\,{\scriptstyle\pm\,2.4\%}$ & $22.8\%\,{\scriptstyle\pm\,1.5\%}$ & $25.3\%\,{\scriptstyle\pm\,0.8\%}$ & $27.8\%\,{\scriptstyle\pm\,1.6\%}$ \\
DeepSeek V3.2 & $20.4\%\,{\scriptstyle\pm\,1.6\%}$ & $24.4\%\,{\scriptstyle\pm\,1.3\%}$ & $24.0\%\,{\scriptstyle\pm\,4.4\%}$ & $33.0\%\,{\scriptstyle\pm\,0.7\%}$ \\
MiniMax M2.5 & $6.3\%\,{\scriptstyle\pm\,1.0\%}$ & $9.9\%\,{\scriptstyle\pm\,0.7\%}$ & $10.3\%\,{\scriptstyle\pm\,5.4\%}$ & $11.2\%\,{\scriptstyle\pm\,0.6\%}$ \\

        \bottomrule
    \end{tabular*}
\end{table}

\clearpage
\section{Evaluation Prompts}
\label{app:prompts}
The following prompts are used across all models and repetitions.
For full-context alignment and selected-region tasks, the system prompt concatenates the shared instructions, the task-specific block, and the final validation instruction, in that order, separated by blank lines.
Astra's opposite-verdict full-context cases use the same full-context alignment prompt.
The context-omitted prompt is given separately in full.
The shared instructions also mention a defect-only alignment mode, which this paper does not evaluate.
Label 2 (\texttt{not\_decidable}) in the prompts is the undecidable label $\bot$.

\begingroup
\lstdefinestyle{evaluationPrompt}{
    language={},
    basicstyle=\small\ttfamily,
    columns=fullflexible,
    keepspaces=true,
    showstringspaces=false,
    breaklines=true,
    breakatwhitespace=true,
    breakindent=0pt,
    tabsize=2,
    frame=single,
    rulecolor=\color{black!25},
    framerule=.3pt,
    framesep=4pt,
    xleftmargin=5pt,
    xrightmargin=5pt,
    aboveskip=6pt,
    belowskip=6pt
}

\subsection{Shared Full-context Instructions}
\begin{lstlisting}[style=evaluationPrompt]
Assess whether the supplied Lean formalization faithfully represents the natural-language specification, interpreted using its complete source context, formal context, and declared implicit assumptions. All supplied documents are available for every task.

Preserve the stated mathematical objects, hypotheses, constructions, and relationships. Faithfulness concerns representation, independently of whether the statement is true or its proof succeeds.

Use label 1 for a faithful correspondence or an acceptable unmatched mathematical contribution. Use label 0 for a faithfulness defect, including a missing required contribution or an unsupported added restriction.

For alignment tasks, first group coherent mathematical contributions in the two targets and match direct correspondences. Judge these directly; refine a group when it has clearly separable faithful and defective contributions. Then use context for unmatched contributions or interpretations that require it. Read the actual supplied definitions when their meaning matters.

Keep shared object introductions together and complete inference rules together. Avoid independent links for every repeated variable use. Separate setup, hypotheses, and conclusion when they are clearly distinct contributions.

A link relates one natural-language contribution to one formal contribution. Each endpoint selects exactly one kind of region: target, context, or assumption. Several discontinuous pieces may jointly express one contribution, including pieces from multiple context documents of the same side. Never mix target, context, and assumption kinds within one endpoint. An assumption endpoint selects exactly one whole assumption document. Either endpoint may be null, but both cannot be null.

Select precise context passages when they supply the contribution being assessed or establish an interpretation required for that link. Supporting reading material may remain unselected. A required source-context condition or assumption missing from the formalization is linked to a null formal endpoint with label 0. Do not select broad context merely because it is available.

Full alignment includes meaningful faithful correspondences and meaningful unmatched mathematical contributions, with an empty counterpart and their own label. Preserve meaningful binders, hypotheses, definitions, typeclass inputs, and logical relationships. An acceptable omission or addition has label 1. Defect-only alignment includes only label 0 links.

Use selective annotation. Presentation-only material, whitespace, noncontributing punctuation, import and scope commands, routine theorem titles, and proof placeholders may remain unselected. Do not add links just to cover characters. Meaningful extras still belong in full alignment.

Spans use zero-based Unicode codepoint offsets and half-open intervals [start,end). Use find_text to obtain exact spans. Before submitting, call validate_answer with the complete proposed JSON string. Correct structural or coordinate errors and validate again. Validation applies from the first attempt and provides no mathematical correctness feedback.

Return only the task's JSON fields, with integer labels 0 or 1, no explanations, and no Markdown fences.
\end{lstlisting}

\noindent\textbf{Final validation instruction.}
This block follows the task-specific instructions.
\begin{lstlisting}[style=evaluationPrompt]
Before submitting your final answer, call validate_answer with the complete proposed JSON string. Correct errors and validate again. Submit the checked JSON as the entire final response. Validation checks structure and coordinates only; you decide the mathematical labels and links.
\end{lstlisting}

\subsection{Full-context Alignment}
\begin{lstlisting}[style=evaluationPrompt]
Construct the complete alignment and assign a binary label to every link. Include meaningful faithful correspondences, defects, and meaningful unmatched mathematical contributions with either label. Return all relevant links without truncation.

Return exactly this JSON structure:
{"findings":[{"N":[["N0",0,10]],"L":[["L0",20,30]],"y":1}]}

N selects natural-language target, source-context, or assumption documents. L selects formal-target or formal-context documents. Each endpoint is a nonempty list of [document_id,start,end] pieces of one kind, or null. Meaningful unmatched NL content has L=null; meaningful unmatched formal content has N=null. Label acceptable omissions or additions 1 and defects 0.
\end{lstlisting}

\subsection{Selected-region Counterpart Identification}
\noindent\textbf{Informal-target regions.}
\begin{lstlisting}[style=evaluationPrompt]
For every supplied natural-language target anchor, find all relevant counterparts and label each link. An anchor's spans form one selection; text lists the exact selected text of each piece. Use the full documents to interpret it. Counterparts may select the formal target or formal context, or be null for an empty counterpart.

The same supplied span may participate in more than one link, serving different mathematical purposes. Return every relevant counterpart, with a separate label for each link.

Return exactly this JSON structure:
{"findings":[{"anchor_id":"a0","other":[["L0",0,10]],"y":1}]}

Every supplied anchor ID must appear at least once. Multiple counterparts and different labels for the same anchor are allowed. Each other endpoint is a nonempty list of [document_id,start,end] pieces of one kind on the opposite side, or null. Return only links incident to supplied anchors. If there are no anchors, return {"findings":[]}.
\end{lstlisting}

\noindent\textbf{Formal-target regions.}
\begin{lstlisting}[style=evaluationPrompt]
For every supplied Lean target anchor, find all relevant counterparts and label each link. An anchor's spans form one selection; text lists the exact selected text of each piece. Use the full documents to interpret it. Counterparts may select the natural-language target, source context, or eligible whole assumption, or be null for an empty counterpart.

The same supplied span may participate in more than one link, serving different mathematical purposes. Return every relevant counterpart, with a separate label for each link.

Return exactly this JSON structure:
{"findings":[{"anchor_id":"a0","other":[["N0",0,10]],"y":1}]}

Every supplied anchor ID must appear at least once. Multiple counterparts and different labels for the same anchor are allowed. Each other endpoint is a nonempty list of [document_id,start,end] pieces of one kind on the opposite side, or null. Return only links incident to supplied anchors. If there are no anchors, return {"findings":[]}.
\end{lstlisting}

\subsection{Context-omitted Alignment}
\begin{lstlisting}[style=evaluationPrompt]
Assess whether the supplied Lean formalization faithfully represents the natural-language specification.

Documents with side N give the natural-language target and any supplied source context or assumptions. Documents with side L give the formal target and formal context. Use all supplied documents and ordinary mathematical and Lean knowledge. Read the definitions used by the formal target when their content affects the comparison.

Faithfulness concerns whether the objects, hypotheses, constructions, and relationships are represented correctly, independently of whether the mathematical statement is true or its proof succeeds.

Use these verdict definitions:
- faithful: the supplied information establishes a faithful representation of the specification.
- unfaithful: the supplied information establishes a faithfulness defect, such as a missing requirement, an unsupported restriction, or a mismatch in a used definition.
- not_decidable: the supplied source specification leaves a substantive interpretive choice unresolved, and reasonable interpretations consistent with the visible wording yield different faithfulness judgments for the fixed formalization.

Use not_decidable for insufficient specification, not for unfamiliar syntax, lack of confidence, or an unproved theorem. Do not invent arbitrary redefinitions or vacuous assumptions. An established defect makes the whole item unfaithful even if another part has an unresolved interpretation.

Construct a complete alignment between the supplied natural-language specification and Lean formalization. Apply the verdict definitions to each mathematical contribution, assigning one label per link: 1 for faithful, 0 for unfaithful, and 2 for not_decidable. An acceptable unmatched contribution receives label 1.

Label every distinct source of undecidability that you identify, while also aligning contributions whose faithfulness is decidable.

First match coherent mathematical contributions in the two targets. Keep shared object introductions and complete inference rules together. Separate setup, hypotheses, defining conditions, and conclusions when they express distinct contributions. Refine a group when clearly separable contributions have different labels. Avoid separate links for every repeated variable use.

When a contribution depends on a formal definition, inspect that definition and select the precise formal passage that explains the correspondence, defect, or unresolved interpretation. For label 2, select the visible source phrase whose interpretation is unresolved and the formal contribution whose faithfulness depends on it. If an endpoint has no corresponding visible contribution, use null. Do not flag an entire theorem or definition when a more precise range identifies the issue.

Each N endpoint is a nonempty list of natural-language spans or null. Each L endpoint is a nonempty list of formal-target spans or formal-context spans, or null; do not mix target and context within one endpoint. Discontinuous pieces may jointly express one contribution. Both endpoints cannot be null. Include meaningful unmatched mathematical contributions with their own label. Routine names, imports, scope commands, punctuation, and proof placeholders may remain unselected.

Spans use zero-based Unicode codepoint offsets and half-open intervals [start,end). Use find_text to locate exact spans in the supplied documents. Before submitting, call validate_answer with the complete proposed JSON string and correct structural or coordinate errors. This tool gives no mathematical correctness feedback.

Return only this JSON structure, with integer labels 0, 1, or 2 and no explanations or Markdown fences:
{"findings":[{"N":[["N0",0,10]],"L":[["L0",20,30]],"y":1}]}
\end{lstlisting}

\subsection{User Message}
The user message is the item's document packet serialized as JSON with two-space indentation.
For full-context and context-omitted alignment, its top-level field is \texttt{documents}.
Each document supplies its \texttt{id}, \texttt{side} (\texttt{N} or \texttt{L}), \texttt{role} (\texttt{target}, \texttt{context}, or \texttt{assumption}), and complete \texttt{text}.
Context-omitted packets contain the two targets and formal context.
Selected-region tasks additionally supply an \texttt{anchors} array; each anchor has an \texttt{id}, its \texttt{spans}, and a \texttt{text} array containing the selected text of each span.
All spans use \texttt{[document\_id, start, end]} with zero-based Unicode-codepoint offsets and half-open intervals.
Reference counterparts and labels are withheld.
\endgroup

\end{document}